\documentclass{article}

\usepackage{microtype}
\usepackage{graphicx}
\usepackage{subfigure}
\usepackage{booktabs} 
\usepackage{wrapfig}
\usepackage{siunitx}
\usepackage{paralist}
\usepackage{url}
\usepackage{rotating}

\usepackage{amsmath,amsfonts,bm}

\def\eqref#1{equation~(\ref{#1})}

\def\1{\bm{1}}

\def\vk{{\bm{k}}}

\def\vt{{\bm{t}}}

\def\vv{{\bm{v}}}
\def\vw{{\bm{w}}}
\def\vx{{\bm{x}}}
\def\vy{{\bm{y}}}

\def\mW{{\bm{W}}}

\DeclareMathAlphabet{\mathsfit}{\encodingdefault}{\sfdefault}{m}{sl}
\SetMathAlphabet{\mathsfit}{bold}{\encodingdefault}{\sfdefault}{bx}{n}

\def\sC{{\mathbb{C}}}

\def\sN{{\mathbb{N}}}

\def\sR{{\mathbb{R}}}

\def\sZ{{\mathbb{Z}}}

\usepackage{hyperref}

\newcommand{\downmapsto}{\rotatebox[origin=c]{-90}{$\scriptstyle\longmapsto$}\mkern2mu}

\usepackage[accepted]{icml2023}

\usepackage{amsmath}
\usepackage{amssymb}
\usepackage{mathtools}
\usepackage{amsthm}

\usepackage[capitalize,noabbrev]{cleveref}

\theoremstyle{plain}

\theoremstyle{definition}

\theoremstyle{remark}

\usepackage[textsize=tiny]{todonotes}
\usepackage{cleveref}
\crefname{section}{Sec.}{Sec.}
\crefname{appendix}{App.}{App.}
\crefname{definition}{Def.}{Defs.}
\crefname{proposition}{Prop.}{Props.}
\crefname{equation}{Eq.}{Eqs.}
\crefname{table}{Tab.}{Tabs.}
\crefname{figure}{Fig.}{Figs.}

\icmltitlerunning{Ewald-based Long-Range Message Passing for Molecular Graphs}

\begin{document}

\twocolumn[
\icmltitle{Ewald-based Long-Range Message Passing for Molecular Graphs}
%



\icmlsetsymbol{equal}{*}

\begin{icmlauthorlist}
\icmlauthor{Arthur Kosmala}{mpl,tucomp,lmuphys}
\icmlauthor{Johannes Gasteiger}{google,tucomp}
\icmlauthor{Nicholas Gao}{tucomp}
\icmlauthor{Stephan Günnemann}{tucomp}
\end{icmlauthorlist}

\icmlaffiliation{mpl}{Max Planck Institute for the Science of Light \& Friedrich-Alexander-Universität Erlangen-Nürnberg, Germany}
\icmlaffiliation{lmuphys}{Department of Physics, Ludwig-Maximilians-Universität München, Germany}
\icmlaffiliation{google}{Google Research}
\icmlaffiliation{tucomp}{Department of Computer Science \& Munich Data Science Institute, Technical University of Munich, Germany}

\icmlcorrespondingauthor{Arthur Kosmala}{arthur.kosmala@tum.de}

\icmlkeywords{Machine Learning, ICML}

\vskip 0.3in
]



\printAffiliationsAndNotice{}  

\begin{abstract}
Neural architectures that learn potential energy surfaces from molecular data have undergone fast improvement in recent years. A key driver of this success is the Message Passing Neural Network (MPNN) paradigm. Its favorable scaling with system size partly relies upon a spatial distance limit on messages. While this focus on locality is a useful inductive bias, it also impedes the learning of long-range interactions such as electrostatics and van der Waals forces. To address this drawback, we propose Ewald message passing: a nonlocal Fourier space scheme which limits interactions via a cutoff on frequency instead of distance, and is theoretically well-founded in the Ewald summation method. It can serve as an augmentation on top of existing MPNN architectures as it is computationally inexpensive and agnostic to architectural details. We test the approach with four baseline models and two datasets containing diverse periodic (OC20) and aperiodic structures (OE62). We observe robust improvements in energy mean absolute errors across all models and datasets, averaging \SI{10}{\percent} on OC20 and \SI{16}{\percent} on OE62. Our analysis shows an outsize impact of these improvements on structures with high long-range contributions to the ground truth energy.
\end{abstract}

\section{Introduction}
\label{intro}

Graph neural networks (GNNs) have shown great promise in learning molecular properties from quantum chemical reference data, speeding up inference times by several orders of magnitude \citep{gilmer_neural_2017} while often preserving the accuracy of their ground truth method. Unfortunately, despite recent success in understanding theoretically universal models \citep{gasteiger_gemnet_2021, dym_universality_2021}, computational constraints still hamper GNNs in learning long-ranged physical interactions between atoms: the standard notion of a molecular graph (if no hand-crafted bond information is used) treats atoms as vertices and connects them by an edge if their 3D distance is within a cutoff hyperparameter \citep{schutt_schnet_2018}. This can be a problem if more remote atom interactions contribute to a target: they are long-ranged also on the graph, potentially relating vertices many edges apart. There is ample empirical and theoretical evidence of limited GNN performance in this case, e.g., due to over-squashing \citep{alon_bottleneck_2021}.\par

\begin{figure}[t]
\includegraphics[width=\linewidth]{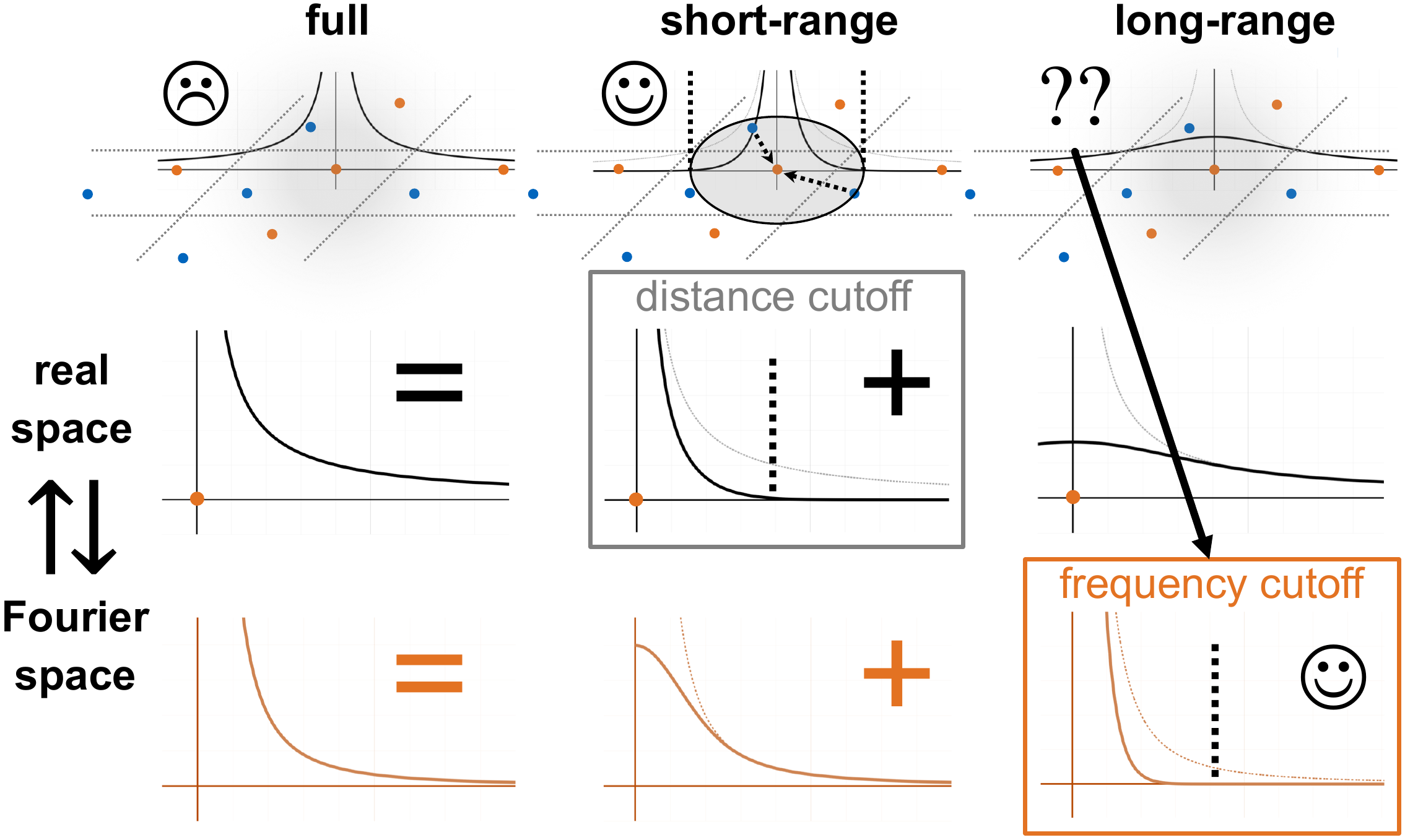}
\vskip -0.02in
\caption{Ewald decomposition of an interaction,
exhibiting $\lVert \vx_i - \vx_j \rVert^{-1}$ decay in this example, into short- and long-range parts.}
\label{fig:ewald_concept}
\vskip -0.2in
\end{figure}

The GNN \emph{status quo} is in sharp contrast to the reality in molecular dynamics, where long-range interactions are an elementary ingredient. A task at the heart of the field (and of this work in particular) is to predict the potential energy $E(\vx_1, \dots, \vx_N)$ based on the atom positions $\{\vx_1, \dots, \vx_N\}$ in a molecule. Hand-crafted parametrizations of this map, known as empirical force fields, distinguish between bonded and non-bonded energy terms, $E = E^{\text{b}} + E^{\text{nb}}$  \citep{leach_molecular_2001}. The bonded terms model the short-ranged nature of covalent bonding with its complex, but highly local interactions. The non-bonded terms, however, capture interactions with heavy-tailed (typically power-law) decay in distance. An example of such a term is the electrostatic interaction,
\begin{equation}
E^{\text{es}} = \sum_{i<j} q_i q_j \lVert \vx_i - \vx_j \rVert^{-1},
\label{eq:esenergyraw}
\end{equation}
where the partial charges $q_i$ quantify net electric charge around atom $i$. Their attraction or repulsion decays inversely with distance, and the energy is the sum of all atom pair interactions. Due to its scale-free decay, simply truncating such a power law with a distance cutoff can induce severe artifacts in measurable thermochemical predictions, even at outsized cutoffs \citep{ewald_mol_sim}. Instead, the interactions of all atom pairs have to be included in the sum. The technique of \emph{Ewald summation} achieves this efficiently. It decomposes the actual physical interaction into a short-range and long-range part. The short-range part can be summed directly with a distance cutoff; the long-range part, while slowly decaying in distance, has a quickly decaying Fourier transform and can be efficiently evaluated as a Fourier space sum with a frequency cutoff (cf. \cref{fig:ewald_concept}).\par

This work lays out a bridge between Ewald summation and the current landscape of GNN models. Our proposed method of \emph{Ewald message passing} (Ewald MP) is a general framework that complements existing GNN layers in analogy to how the frequency-truncated long-range part complements the distance-truncated short-range part in Ewald summation. It is architecture-agnostic and computationally efficient, which we demonstrate by implementing and testing it as a modification on top of existing GNN models. Our experiments on the periodic structures of the OC20 dataset and the aperiodic structures of the OE62 dataset indicate robust improvements in energy mean absolute errors averaging \SI{10}{\percent} and \SI{16}{\percent} OC20 and OE62, respectively. Moreover, we study the tradeoff of these improvements against runtime and parameter count and find that it cannot be consistently replicated by any tested baseline setting. In a closer examination of the OE62 results, we observe that Ewald message passing recovers the effects of DFT-D3, a hand-crafted long-range correction term. This supports our claim that the method targets the learning of long-range interactions in particular. In summary, our contributions are:
\begin{compactitem}
\item The new framework of \textbf{Ewald message passing}, which is readily integrated with existing GNN models.
\item Insights abouts its \textbf{cost-performance} tradeoff in comparison to the wider GNN design landscape.
\item A study underscoring the utility of Ewald message passing as a \textbf{long-range correction scheme}.
\end{compactitem}

\section{Background}
\paragraph{Message passing neural networks.}
MPNNs were introduced by \citet{gilmer_neural_2017} as a conceptual framework covering many molecular GNN architectures. Molecules are represented as graphs with atoms $i$ as vertices. Atoms are adjacent, $j \in \mathcal{N}(i)$, if $\lVert \vx_i - \vx_j\rVert < c_x$, where $c_x$ is the distance cutoff. An MPNN first featurizes atoms as embeddings $h_i^0 \in \sR^F$ based on local atom properties alone. Next, it models the complex atomic interactions by iteratively updating the embeddings in \emph{message} and \emph{update} steps,
\begin{align}
M_i^{(l+1)} &= \sum_{j \in \mathcal{N}(i)} f_\text{int}\left(h^{(l)}_i, h^{(l)}_j, e_{(ij)}\right),\label{eq:messagepassing}\\
h_i^{(l+1)} &= f_{\text{upd}}\left(h_i^{(l)}, M_i^{(l+1)}\right),
\end{align}
where the \emph{message sum} $M_i^{(l+1)}$ gathers information from the neighborhood $\mathcal{N}(i)$, using a function $f_\text{int}$ that depends on current embeddings and edge features $e_{(ij)}$ (e.g., pairwise distances). Next, a learnable function $f_\text{update}$ processes the message sums. A \emph{readout step} then generates outputs (e.g., atom-wise energy contributions) from the embeddings. We develop the core framework of Ewald message passing around a particular type of message step, the continuous-filter convolution \citep{schutt_schnet_2018}:
\begin{equation}
    M_i^{(l+1)} = \sum_{j \, \in \, \mathcal{N}(i)}{h_j^{(l)}} \cdot \Phi^{(l)}(\lVert\vx_i - \vx_j\rVert),
    \label{eq:cfconv}
\end{equation}
where $\cdot$ acts component-wise in the feature dimension, and $\Phi^{(l)}(\lVert\cdot\rVert) \colon \sR^3 \longrightarrow \sR^F$ is a stack of $F$ learned, radial filters. In every channel $1 \leq d \leq F$, $\Phi^{(l)}_d(\lVert\cdot\rVert)$ learns attending to fuzzy distance ranges around an atom, allowing regression tasks on continuously varying molecular geometries.

\paragraph{Periodic systems.}
\begin{wrapfigure}[12]{r}{0.5\linewidth}
\begin{center}
\includegraphics[width=\linewidth]{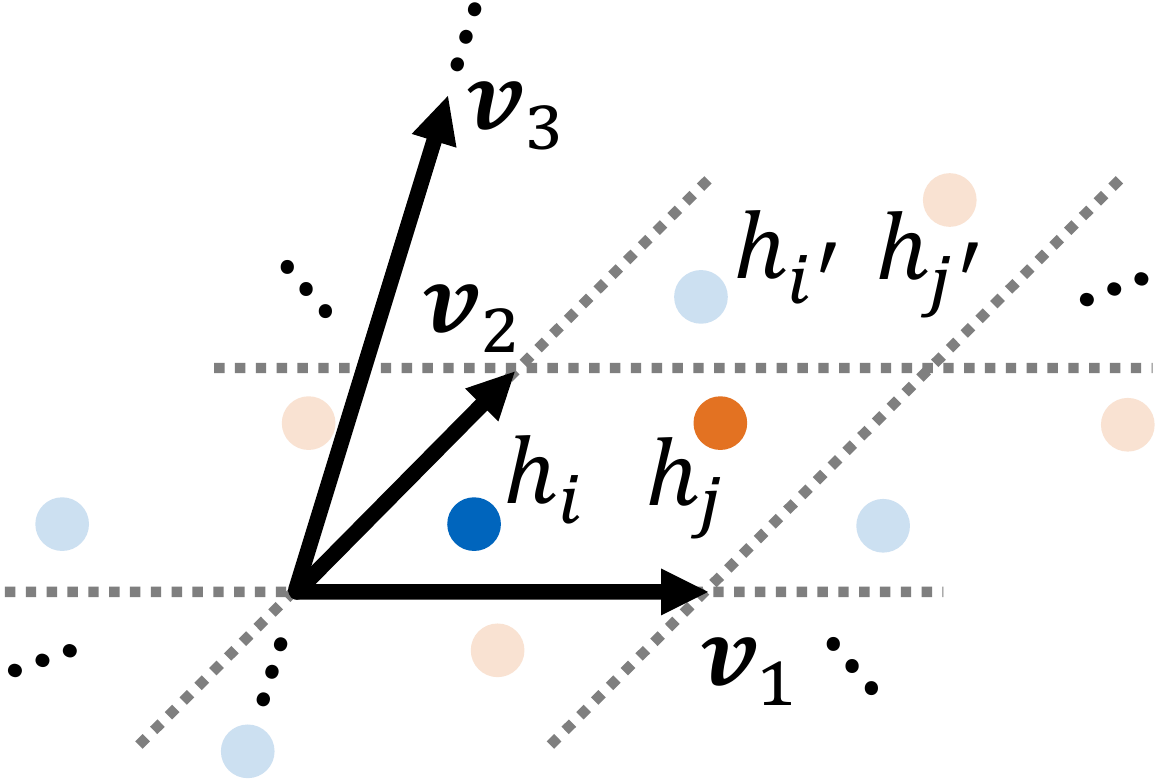}
\vskip -0.02in
\caption{Schematic illustration of periodic boundary conditions.}
\label{fig:ewald1}
\end{center}
\end{wrapfigure}
The inadequacy of a distance cutoff for generic interactions becomes particularly apparent in materials, where interatomic distances are unbounded.\par
\emph{Periodic boundary conditions} (PBC) can be used to approximate an infinite system as repeating copies of a finite atom collection (cf. \cref{fig:ewald1}), tiling up space within \emph{supercells}\footnote{A supercell should not be confused with the unit cell of a crystalline material. Unit cell translation symmetry is broken in non-equilibrium structures, but PBC are still applied, with a typical supercell containing multiple unit cells.} spanned by the lattice vectors $(\vv_1, \vv_2, \vv_3) \in \sR^{3 \times 3}$. Let $\mathcal{S} = \{(h_1, \vx_1), \dots (h_N, \vx_N)\}$ denote the atom embeddings and positions of one chosen supercell and let $\vt \in \Lambda$ be any integer shift on the supercell lattice $\Lambda \vcentcolon = \{\lambda_1 \vv_1 + \lambda_2 \vv_2 + \lambda_3 \vv_3 \mid \boldsymbol{\lambda} \in \sZ^3\}$. Atoms $i'$ at $\vx_{i'} \vcentcolon = \vx_i + \vt$ are called \emph{images} of $i \in \mathcal{S}$, with the PBC fixing $h_{i'} \equiv h_i$. \par
Under PBC, generic message sums become infinite series as every image may contribute a message. Previous MPNN work on materials mitigates this issue with a distance cutoff, alongside a multigraph representation that avoids storing redundant $h_{i'}$ on separate vertices \citep{xie_crystal_2018, chanussot_open_2021, Cheng2021, xie_crystal_diffusion}.


\section{Ewald Message Passing}
\label{sec:EwaldMP}

\paragraph{Analogy to electrostatics.}
The reason why Ewald summation can be applied to MPNNs is a formal correspondence between electrostatics-type interactions and continuous-filter convolutions. Previous work has highlighted this analogy as a useful source of inductive bias \citep{anderson_cormorant:_2019}. The connection becomes clear if we take \cref{eq:esenergyraw} for the electrostatic energy $E^{\text{es}}$ and express it in a new notation alluding to the continuous-filter convolution of \cref{eq:cfconv}:
\begin{align}
&E^{\text{es}} = \frac{1}{2}\sum_{i=1}^N{q_i \cdot V_i^{\text{es}}(\vx_i)},\\
&V_i^{\text{es}}(\vx_i) = \sum_{j \neq i} q_j \cdot \Phi^{\text{es}}(\lVert \vx_i - \vx_j \rVert),\quad
\Phi^{\text{es}}(r) = \frac{1}{r},
\label{eq:esenergy}
\end{align}
where $N$ is the total number of atoms ($N \vcentcolon = \infty$ in materials) and we introduced the \emph{electric potentials} $\{V_i^{\text{es}} \mid i = 1, \dots, N\}$ and the \emph{interaction kernel} $\Phi^{\text{es}}$. Compare the expression for $V_i^{\text{es}}$ to the message sum  $M_i^{(l+1)}$ in \cref{eq:cfconv}: setting the cutoff $c_x \rightarrow \infty$ (such that $\mathcal{N}(i) = \{1, \dots, N \} \setminus \{i\}$) and substituting
\begin{equation}
q_j \leftrightarrow h_j^{(l)}, \quad \Phi^{\text{es}} \leftrightarrow \Phi^{(l)}, \quad V_i^{\text{es}}(\vx_i) \longleftrightarrow M_i^{(l+1)}
\end{equation}
puts $F$-dimensional embedding vectors in place of scalar-valued partial charges as atom descriptors, and replaces the simple, domain-informed $1/r$ kernel by a stack of $F$ learned convolution filters formally acting as interaction kernels. Any continuous-filter message sum $M_i$ (we drop the layer superscript $(l)$ from now on) can be \emph{canonically identified} with a potential function $V_i(\vx_i)$ evaluated at source atom location $\vx_i$. Aggregating atomic output after just one message-passing step would inductively bias the model towards two-body interaction energies of similar form to $E^{\text{es}}$. Several message and update steps in principle enable the modeling of $n$-body terms \citep{batatia_design_space}.

\begin{figure*}[t!]
\includegraphics[width=\linewidth]{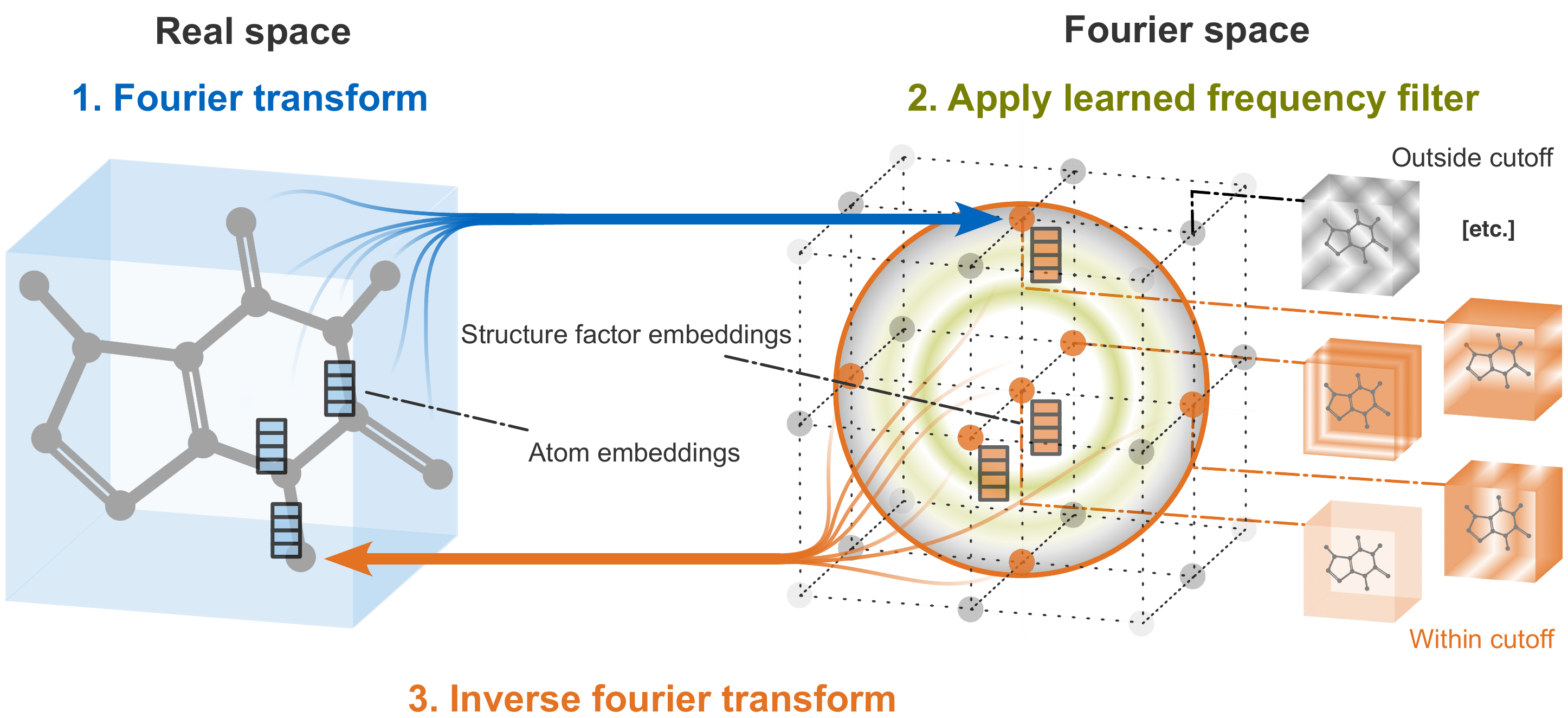}
\vskip -0.001in
\caption{Three-part structure of long-range message passing: first, structure factor embeddings $\{s_{\vk}\}$ are evaluated as a Fourier sum over atoms $\sum_{j \in \mathcal{S}} \exp(-i\vk\vx_{j}) \, [\, \cdot \,]$ and stored. Next, these are multiplied $\vk$-wise with learned filter values. Finally, an inverse Fourier sum $\sum_{\vk} \exp(i\vk\vx_{i}) [\, \cdot \,]$ scatters messages back to each atom $i$. Note the nonlocal \emph{all-to-one} character of this scheme. If PBC are explicitly prescribed, the Fourier domain is given by $\{\vk \in \Lambda' \mid \lVert \vk \rVert < c_k\}$ with frequency cutoff $c_k$. If not, it needs to be suitably discretized.}
\label{fig:ewald_main}
\end{figure*}

\paragraph{Conceptual outline.}
We now give an overview of Ewald summation and show how it connects to our learning approach, Ewald message passing, based on the above analogy.\par
\emph{Ewald summation} evaluates potentials (equivalently, message sums) $V_i(\vx_i) = M_i$ given a PBC lattice, a generic interaction kernel $\Phi \colon \sR^+ \rightarrow \sR^F$ and generic atom ``charges'' $h_i \in \sR^F$.
We apply Ewald summation if $\Phi$ is long-ranged.\par In this case, the naive distance cutoff approximation $M_i \approx \sum_{j \neq i, \lVert \vx_i - \vx_j \rVert < c_x} h_j \cdot \Phi(\lVert \vx_i - \vx_j \rVert)$ becomes inaccurate for practically-sized values of $c_x$.
A slow distance decay of $\Phi$ means that a significant part of its tail remains outside of $c_x$, even for large $c_x$. This long-range information is subsequently lost by truncation. Ewald summation presumes a known decomposition $\Phi(r) = \Phi^{\text{sr}}(r) + \Phi^{\text{lr}}(r)$ into a \emph{short-range} and \emph{long-range} kernel, where
\begin{itemize}
    \item{$\Phi^{\text{sr}}$ has fast decay in distance,}
    \item{$\Phi^{\text{lr}}(0)$ is well-defined, and $\hat\Phi^{\text{lr}}(\lVert \cdot \rVert)$, the 3D Fourier transform of $\Phi^{\text{lr}}(\lVert \cdot \rVert)$, has fast decay in frequency.}
\end{itemize}

By construction, atom sums over distance-truncated short-range kernels $\Phi^{\text{sr}}_{[0,c_x]} \vcentcolon = \Phi^{\text{sr}} \cdot \mathbf{1}_{[0,c_x]}$ (where $\mathbf{1}_{[0,c_x]}$ is the indicator function on $[0,c_x]$) converge quickly for growing $c_x$, meaning that a distance cutoff can be efficiently applied. This cannot be the case for $\Phi^{\text{lr}}$, the leftover fat tail of $\Phi$. Ewald summation deals with this \emph{long-range sum} over $\Phi^{\text{lr}}$ by rewriting it as a sum over the Fourier transform $\hat \Phi^{\text{lr}}$. Replacing $\hat \Phi^{\text{lr}}$ by $\hat \Phi^{\text{lr}}_{[0,c_k]} \vcentcolon = \hat\Phi^{\text{lr}} \cdot \mathbf{1}_{[0,c_k]}$, i.e., truncation by a \emph{frequency cutoff} $c_k$, now admits efficient evaluation of the long-range sum. Adding the short-range and long-range sums recovers the sum over the full kernel $\Phi$. \cref{fig:ewald_concept} summarizes the approach. Convenient decompositions satisfying the above properties for all power law kernels are provided by \citet{nijboer_lattice_sums}.\par

\emph{Ewald message passing} essentially flips the rationale behind Ewald summation: instead of starting with a known physical kernel $\Phi$ and searching for a decomposition with the above properties, we seek to \emph{parametrize} a filter $\Phi$ that is not short-ranged. However, we impose that it can be fitted by an Ewald sum over cutoff-supported short-range and long-range kernels. The kernels are learned independently of each other, $\hat \Phi^{\text{lr}}_{[0,c_k]}$ is learned directly in Fourier space.\par

This section proceeds by rewriting the crucial long-range sum as a sum over Fourier frequencies, bringing it into a form that admits a frequency cutoff. This prepares a subsequent in-depth discussion of Ewald message passing and its implementation as part of existing GNN models.

\paragraph{Frequency truncation.}
\begin{wrapfigure}[14]{r}{0.5\linewidth}
\begin{center}
\includegraphics[width=\linewidth]{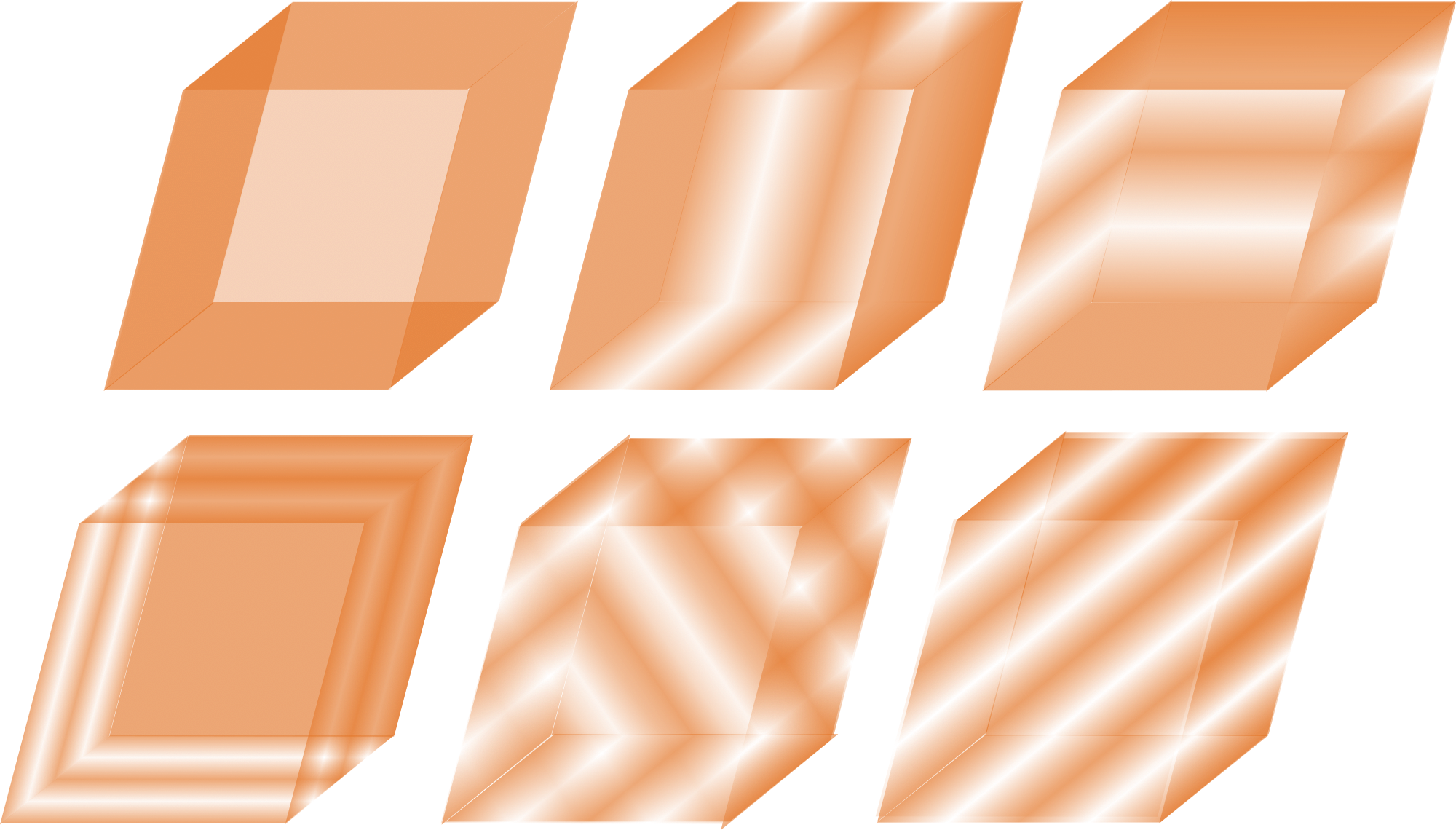}
\caption{Sketched surfaces of constant phase for Fourier modes $\exp(i \vk^T \vx)$ at small $\vk \in \Lambda'$.}
\label{fig:fouriermodes}
\end{center}
\end{wrapfigure}
From here on, we assume that the long-range sum runs over \emph{all} atoms, \emph{including} target atom $i,$ and thereby denote it by $M^{\text{lr}}(\vx_i)$ (without subscript-$i$, but highlighting the dependence on source atom position $\vx_i$). Note that this minor redefinition turns the map $\vx \mapsto M^{\text{lr}}(\vx)$ into a periodic function. We check this by phrasing it as a double sum over the atoms of one supercell $\mathcal{S}$ and the supercell lattice $\Lambda$:
\begin{align}
\begin{split}
\label{eq:periodic}
M^{\text{lr}}(\vx) &= \sum_{\vt \in \Lambda}\sum_{j \in \mathcal{S}} h_j \cdot \Phi^{\text{lr}}(\lVert\vx - (\vx_j + \vt)\rVert)\\
&= M^{\text{lr}}(\vx + \tilde \vt), \qquad \forall \tilde \vt \in \Lambda.
\end{split}
\end{align}
Thus, $M^{\text{lr}}$ has a Fourier series expansion (we use complex exponentials as Fourier modes for notational compactness):
\begin{equation}
M^{\text{lr}}(\vx_i) = \sum_{\vk \in \Lambda'} \hat M_\vk^{\text{lr}} \exp{(i \vk^T \vx_i)}
\label{eq:fourierseries}
\end{equation}
with Fourier coefficients $\hat M_\vk^{\text{lr}} \in \sC^F$ (determined next). Just as frequencies of a 1D Fourier series must be integer multiples of the basis frequency (i.e., form a 1D lattice), the 3D spatial frequencies $\vk \in \Lambda' \subset \sR^3$ must be integer combinations of \emph{three} spatial basis frequencies $(\vw_1, \vw_2, \vw_3) \in \sR^{3 \times 3}$ spanning the \emph{reciprocal lattice} $\Lambda' \vcentcolon = \{\lambda'_1 \vw_1 + \lambda'_2 \vw_2 + \lambda'_3 \vw_3 \mid \boldsymbol{\lambda'} \in \sZ^3\}$. Each basis frequency amounts to a single-period phase shift along one supercell direction and constant phase along the others, i.e., $\vw_i^T \vv_j = 1$ if $i=j$ and $0$ otherwise (cf. \cref{fig:fouriermodes}). The supercell lattice $\Lambda$ therefore fixes the reciprocal lattice $\Lambda'$, and the relations $\vw_1 = 2\pi(\vv_2 \times \vv_3)\Omega^{-1}$, $\vw_2 = 2\pi(\vv_3 \times \vv_1)\Omega^{-1}$, $\vw_3 = 2\pi(\vv_1 \times \vv_2)\Omega^{-1}$ (with the cross product $\times$ and supercell volume $\Omega = \vv_1^T (\vv_2 \times \vv_3)$) are easily verified.\par

As our final ingredient leading up to a frequency cutoff, we relate the Fourier \emph{coefficients} $\hat M_\vk^{\text{lr}}$ to the 3D Fourier \emph{transform} $\hat \Phi^{\text{lr}}(\lVert \cdot \rVert) \colon \sR^3 \rightarrow \sR^F$ of $\Phi^{\text{lr}}(\lVert \cdot \rVert)$ – the kernel is aperiodic unlike $M^{\text{lr}}(\cdot)$, implying that its frequency spectrum is a continuous, radial function of $\vk \in \sR^3$ rather than a discrete set of lattice coefficients. In \cref{app:relationship}, we derive the identity $\hat M_\vk^{\text{lr}} = \frac{1}{\Omega} \sum_{j \in \mathcal{S}} h_j \exp(-i\vk^T \vx_j) \hat \Phi^{\text{lr}}(\lVert \vk \rVert)$. Essentially, \cref{eq:periodic} periodizes $\sum_{j \in \mathcal{S}} h_j \cdot \Phi^{\text{lr}}(\lVert \cdot - \vx_j \rVert)$ by adding all of its lattice shifts on $\Lambda$; in Fourier space, this periodization maps $\hat \Phi^{\text{lr}}(\lVert \cdot \rVert)$ onto a discrete spectrum by recording its values only at PBC-compatible frequencies in $\Lambda'$. Inserting this back into \cref{eq:fourierseries}, and absorbing the prefactor $\frac{1}{\Omega}$ in our \emph{learned} filter by redefining $\frac{1}{\Omega} \hat \Phi^{\text{lr}}(\lVert \vk \rVert) \mapsto \hat \Phi^{\text{lr}}(\lVert \vk \rVert)$, yields the desired long-range component of Ewald message passing, a message sum over Fourier frequencies:
\begin{align}
M^{\text{lr}}(\vx_i) &= \sum_{\vk \in \Lambda'} \exp(i \vk^T \vx_i) \cdot s_\vk \cdot \hat \Phi^{\text{lr}}(\lVert \vk \rVert),\label{eq:lrmessage}\\
s_\vk &= \sum_{j \in \mathcal{S}} h_j \exp(-i \vk^T \vx_j),
\label{eq:s}
\end{align}
with $\{s_\vk \in \sC^F \mid \vk \in \Lambda'\}$, which we call \emph{structure factor embeddings}, as inspired by the analogous crystallographic quantity. If we replace $\hat \Phi^{\text{lr}}$ by a cutoff-truncated Fourier space kernel $\hat \Phi^{\text{lr}}_{[0,c_k]}$, we can limit the outer sum $\sum_{\vk \in \Lambda'} \mapsto \sum_{\vk \in \Lambda', \lVert \vk \rVert < c_k}$, while the inner sum computing $s_\vk $ only extends over the atoms of a single supercell. The resulting cutoff-truncated kernel is complementary to regular continuous filters (\cref{eq:cfconv}): While they have no limit on distance, the frequency cutoff does limit how fast the filters can change in space. More specifically, their spatial resolution is proportional to the inverse frequency cutoff.

\paragraph{Structure factor embeddings.} Without a $k$-cutoff, the set of tuples $\{(s_{\vk}, \vk) \mid \vk \in \Lambda'\}$ would hold equivalent information to the set $\{(h_{i}, \vx_{i}) \mid i \in \mathcal{S}\}$, i.e., fully quantify the hidden state of the model on a given input structure. This is seen by viewing $\{(h_{i}, \vx_{i}) \mid i \in \mathcal{S}\}$ equivalently as an ``embedding density distribution'', $\rho(\vx) = \sum_{\vt \in \Lambda}\sum_{i \in \mathcal{S}} h_{i} \delta(\vx - \vx_{i} -\vt)$, with $\delta$ denoting the Dirac delta distribution. Its Fourier transform (in the distributional sense) is given by $\hat \rho(\vk) = \sum_{\vk' \in \Lambda'} \left[\sum_{j \in \mathcal{S}} h_i \exp(-i\vk'^T\vx_j) \right] \delta(\vk - \vk') = \sum_{\vk' \in \Lambda'} s_{\vk'} \delta(\vk - \vk')$ which corresponds to the pairs $\{(s_{\vk}, \vk)\}$.
Disregarding all $s_{\vk}$ outside a $k$-cutoff removes fine-grained resolution information about the atom encoding distribution below a scale $\sim c_k^{-1}$. In this sense, the structure factor representation introduced in Ewald message passing coarse-grains the hidden model state. Moreover, the embedding information is represented non-locally since each $s_{\vk}$ is a weighted sum over \textit{all} supercell embeddings. This all-to-one relationship is another way of understanding the long-range character of Ewald message passing.

\paragraph{Ewald message passing.} The combined usage of regular short-range message passing via \emph{any} GNN layer and the long-range message passing in \cref{eq:lrmessage,eq:s} is called \emph{Ewald message passing}. This combination enables Ewald message passing to express interactions outside of either limited class. We combine the two message passing schemes simply by adding up both embedding updates during each message-passing step, for details see \cref{eq:ewaldmod}. This way of integration works nearly out-of-the-box in our experiments.

\paragraph{Aperiodic systems.}
Many structural datasets containing important long-range interactions do not feature periodic boundary conditions. In this case, a reciprocal lattice that would discretize the Fourier domain is not provided, which in turn means the structure factor embeddings $s_{\vk}$ turn into a function $s(\vk)\colon \sR^3 \mapsto \sC^F$ defined on a continuous domain. The aperiodic case can equally well be understood as the limit of infinite supercell size, corresponding to a continuum limit of the reciprocal lattice as the basis frequency vectors $\vw_i$ approach zero. As we can only represent $s(\cdot)$ and $\hat \Phi(\lVert \cdot \rVert)$ by a finite set of numbers, we need to impose a suitable discretization scheme ourselves. In our implementation, we tile the cutoff region by a grid of cubic voxels and replace all continuous functions by the set of their average values within each voxel. The discretization resolution is controlled by the voxel sidelength $\Delta$, a new hyperparameter that is relevant for the aperiodic case. In our main experiments, we use equal resolution $\Delta = \SI{0.2}{\angstrom}^{-1}$ for each tested model. Ablation studies with other resolutions are reported in \cref{app:ewaldhparams}. Note that this approach preserves rotation invariance only if the voxel grid sits in a coordinate frame that rotates \emph{with} the input structure. To this end, we use the orthonormal basis obtained from a singular value decomposition, similar to \citet{gaoAbInitioPotentialEnergy2022}. In \cref{app:structurefactor} we outline a more general mathematical framework for discretizing the structure factor, provide implementation-relevant details about our voxel approach and discuss its relation to other possible discretization schemes.

\paragraph{Frequency filters: aperiodic case.} Thus far, we left open how the frequency filters $\hat \Phi(\lVert \cdot \rVert)$ are learned. In the aperiodic case, we parametrize filters as linear combinations of $S$ radial basis functions $\{\hat \psi_i(\lVert \cdot \rVert) \colon \sR^3 \rightarrow \sR \mid i = 1, \dots, S\}$: $\hat \Phi(\lVert\vk\rVert) = \mW \hat \Psi(\vk)$ with $\hat\Psi(\vk) = (\psi_1(\lVert\vk\rVert), \dots \psi_S(\lVert\vk\rVert))$ and the learned weight matrix $\mW \in \sR^{F \times S}$. All basis functions are supported within the $k$-cutoff radius $c_k$. We use Gaussian radial basis functions \citep{schutt_schnet_2018}.

\paragraph{Frequency filters: periodic case.} On OC20 we pursue a non-radial filtering strategy that allows for a more liberal use of the information provided by the PBC frame $(\vv_1, \vv_2, \vv_3) \in \sR^{3 \times 3}$ of a given input structure. This information can be meaningful: given a material surface slab, for example, the model should clearly ``know'' which direction is normal to the surface. This does not break rotation invariance as the PBC frame rotates \emph{with} the structure.

Since the reciprocal lattice $\Lambda'$ varies depending on the input PBC frame, the number of frequencies inside any fixed cutoff volume would also be variable. Motivated by our empirical findings (\cref{app:ewaldhparams}), we instead \emph{fix} the number of frequencies in each reciprocal basis direction $(\vw_1, \vw_2, \vw_3) \in \sR^{3 \times 3}$, dropping the notion of a direct frequency cutoff. Formally, we define positive integers $N_x, N_y, N_z$ and consider the product set $I = \bigtimes_{j=x,y,z} \{-N_j, \dots, N_j\} \subset \sZ^3$. We limit the scope of the Fourier message sum in \cref{eq:lrmessage} by only including frequencies with reciprocal lattice positions in $I$. The model then learns a filter weight $\hat \Phi_{\bm{\lambda'}} \in \sR^F$ for each $\bm{\lambda'} \in I$. Since a general supercell lattice has at least point symmetry, we moreover set $\hat \Phi_{\bm{\lambda'}} \equiv \hat \Phi_{\bm{-\lambda'}}$. This differs from the strict notion of frequency filters discussed earlier: the filter weights $\hat \Phi_{\bm{\lambda'}}$ are not assigned to a fixed point $\vk$ in Fourier space, but to a position on the reciprocal lattice. The frequency corresponding to this position $\vk_{\bm \lambda'} = \lambda'_1 \vw_1 + \lambda'_2 \vw_2 + \lambda'_3 \vw_3$ varies with the basis frequencies of each input structure.
Note that this filtering approach is not invariant under the ``supercell symmetry'' that maps any supercell to its multiple across one or several lattice directions while keeping atom positions unchanged. We found that this does not affect empirical performance on OC20. Still, a filtering approach invariant to this symmetry (e.g.\ radial basis filtering) might be advisable for datasets with large variations in supercell size. 

\paragraph{Computational complexity.}
The three-part procedure of our long-range message passing scheme, illustrated in \cref{fig:ewald_main}, has a computational complexity of $\mathcal{O}(N_{\text{at}} N_{\text{k}}$), where $N_{\text{at}}$ is the number of atoms in the structure (or supercell in the case of PBC), and $N_{\text{k}}$ is the number of frequencies contained in the long-range sum. As the density of atoms in space can be roughly assumed constant, the atom count increases linearly with the supercell volume $\Omega$ in the periodic case. By the definition of the reciprocal basis $(\vw_1, \vw_2, \vw_3) \in \sR^{3 \times 3}$, the amount of frequencies $\vk \in \Lambda'$ contained within a fixed frequency cutoff volume therefore scales linearly with the atom count $N_{\text{at}}$. Altogether, this would amount to $\mathcal{O}(N_{\text{at}}^2)$ scaling if both cutoffs are left fixed. In the aperiodic case, the same holds if we adapt the voxel resolution $\Delta$ to the finest features of $s$, which, by Fourier reciprocity, scale inversely to the structural length. However, in contrast to standard Ewald summation \citep{ewald_mol_sim}, Ewald MP does not put any fixed relationship between $c_x$ and $c_k$. To achieve a computational complexity of $\mathcal{O}(N_{\text{at}})$, we can instead choose to fix $N_{\text{k}}$, which effectively varies $c_k \propto N_{\text{at}}^{-1}$. Note that this approach can leave a ``gap'' of medium-range interactions that are covered neither by the short-range nor the long-range component of Ewald MP. Another way to achieve sub-quadratic scaling would be adapting more efficient Ewald summation variants to MPNNs (cf. \cref{sec:rw}).


\section{Related Work} \label{sec:rw}
 
\paragraph{Hand-crafted long-range corrections.}
Approaches augmenting a short-ranged model by a long-range correction make a frequent appearance in computational chemistry. This often concerns corrections to density functional theory (DFT), the method behind the OC20 \citep{chanussot_open_2021}, OE62 \citep{oe62_paper} and many other machine learning benchmarks. DFT suffers from the same type of problem as MPNN potentials: its standard functionals like LDA \citep{lda_paper} or PBE \citep{pbe_paper} cannot correctly express long-range effects such as London dispersion \citep{dft_review}. A zoo of interatomic dispersion corrections exists \citep{dft-d2-paper, ts_correction_paper, dft-d3, dft-d4-paper}, which can be adapted for use alongside many DFT functionals. From this historic angle, Ewald MP fills the same demand for an inexpensive, out-of-the-box correction in the MPNN context. Others have combined long-range terms of various physics-based functional forms with short-range models, including fully hand-crafted terms \citep{staacke_long_range} and electrostatic terms with MPNN-predicted partial charges \citep{unke_spookynet_2021}. In contrast, our approach remains entirely within the MPNN paradigm, focusing on a fully data-driven modeling of long-range effects.

\paragraph{Non-local learning schemes.}
Ewald message passing falls into the broader category of learning approaches targeting non-local interactions in a structure. An example of prominence well beyond quantum chemistry is attention \cite{vaswani_attention_2017}. Whereas early approaches in the graph domain apply the attention mechanism to local node neighborhoods only \citep{velickovic_graph_2018,dwivedi2021generalization}, others soon extended it to the fully-connected graph \citep{ying_transformers_2021,kreuzer2021rethinking,mialon2021graphit}. A drawback of such standard, fully-connected graph attention is its $\mathcal O(N^2)$ scaling in structure size. However, models using subquadratic alterations of standard attention have been proposed, including an $\mathcal{O}(N \log (N))$ non-local aggregation scheme \citep{liu_non_local_gnns}, as well as linearly-scaling modified attention layers for both chemistry \citep{unke_spookynet_2021,frank2022sokrates} and generic graph applications \citep{rampasek_recipe}. Shallow models such as Gaussian-process-based sGDML \citep{chmiela_towards_2018} also typically use locality approximations, motivating recent effort towards their nonlocal extension by \citet{chmiela_accurate}. Such approaches are mostly orthogonal to the MPNN line of research and not directly informed by a computational physics technique as with Ewald MP.

\paragraph{Use of long-range sums.}
Some MPNN-based works share our approach of computing multiple message sums, in analogy to the physical separation of hand-crafted energies into bonded and non-bonded terms. They likewise use a more complex short-range part which includes higher-order geometric information like bond \citep{zhang_molecular_2020} or even dihedral angles \citep{gemnet_oc}, and a simpler, purely radial atom-to-atom long-range part. However, the long-range part in both works is a standard distance-truncated message sum in a larger atomic neighborhood, thereby precluding the scaling advantage and physical inductive bias achievable with Ewald summation. Apart from this, two concurrent works exist which combine forms of Ewald summation with MPNN models. \citet{lin2023efficient} use it in a traditional sense to compute \emph{fixed-form} physical potentials for downstream use as physics-informed edge features, rather than as a \emph{free-form} ansatz for the learning of continuous-filter messages. \citet{yu2022capturing} share our notion of learnable Fourier space filters, but compute a scalar-valued structure factor instead of structure factor embeddings. More generally, \citet{yu2022capturing} do not use the Ewald part as a GNN message passing step with feedback onto the base GNN embeddings, but rather apply it once and then combine it with the short-range GNN output after its final layer. Both Ewald-related works exclusively consider the periodic case.

\paragraph{Ewald summation.}
Ewald summation is a technique for the summation of long-range interactions. The concept dates back to work in theoretical physics from over a century ago \citep{ewald_original_paper} and has been adapted various times to encompass more general interactions \citep{nijboer_lattice_sums, ewald_multipole_paper, ewald_2d_paper}. Further development has focused on the numerically efficient evaluation of the long-range sum. While direct Ewald summation with optimized distance and frequency cutoffs has $N^{\frac{3}{2}}$ scaling in the atom number \citep{ewald_mol_sim}, approaches using the Fast Fourier Transform \citep{particle_mesh_ewald, fast_fourier_poisson_ewald, smooth_particle_mesh_ewald} bring it down to $N \log(N)$. In this work, we focus on introducing the general method of Ewald message passing, and consider numerical improvements an interesting route for future work.

\section{Experiments}
\label{sec:exp}

\paragraph{Datasets.} The OC20 dataset \citep{chanussot_open_2021} features adsorption energies and atom forces for roughly 265 million structures, each one a snapshot from a DFT-computed relaxation trajectory of an adsorbate-catalyst combination under 3D periodic boundary conditions. We train our models on the OC20-2M subsplit, which has been empirically established as an efficient proxy for the full-size training set \citep{gemnet_oc}. The OE62 dataset \citep{oe62_paper} features, among other targets, DFT-computed energies (in \si{\electronvolt}) for roughly 62,000 large organic molecules. The energies account for the long-ranged London dispersion interaction, computed via the DFT-D3 dispersion correction \citep{dft-d3}. Both benchmarks contain structures with comparably large numbers of atoms ($N>100$). OC20 features PBC, while structures in OE62 are aperiodic but can reach large spatial extent ($>\SI{20}{\angstrom}$). This makes these two datasets a strong testing case for the long-range improvements achievable with Ewald message passing.

\paragraph{Baseline GNN models.} An overview of training and baseline hyperparameters can be found in \cref{app:training}. We test Ewald message passing by modifying four models: SchNet \citep{schutt_schnet:_2017}, PaiNN \citep{painn_schutt}, DimeNet$^{++}$ \citep{gasteiger_fast_2020} and GemNet-T \citep{gasteiger_gemnet_2021}. These are intended as a representative, but by no means exhaustive sample from the modern GNN landscape. Guided by the settings used by \citet{chanussot_open_2021}, most baseline atom embedding sizes and cutoffs are larger than the originally reported values to account for the size and chemical diversity of our structures.

\begin{figure*}[t]
\begin{minipage}[t]{\columnwidth}
\includegraphics[width=\linewidth]{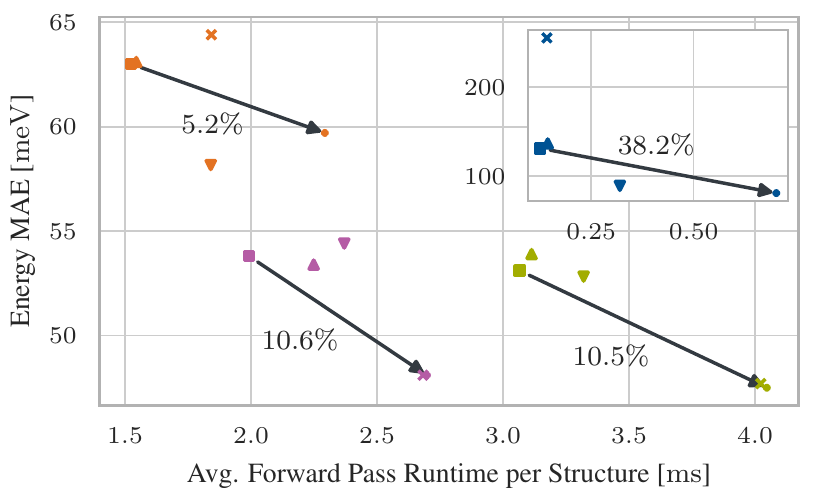}
\vskip -0.05in
\caption{Cost vs.\ energy MAE on OE62. Ewald MP improves the models by \SIrange{5}{38}{\percent}. Increasing the cutoff performs similarly for DimeNet$^{++}$ and GemNet-T, but is even detrimental for SchNet and PaiNN. Using a SchNet in place of an Ewald LR block works well for SchNet and PaiNN, but leaves DimeNet$^{++}$ and GemNet-T unaffected. Increasing the atom embedding size has little effect.}
\label{fig:pareto_oe62}
\end{minipage}
\hfill
\begin{minipage}[t]{\columnwidth}
\includegraphics[width=\linewidth]{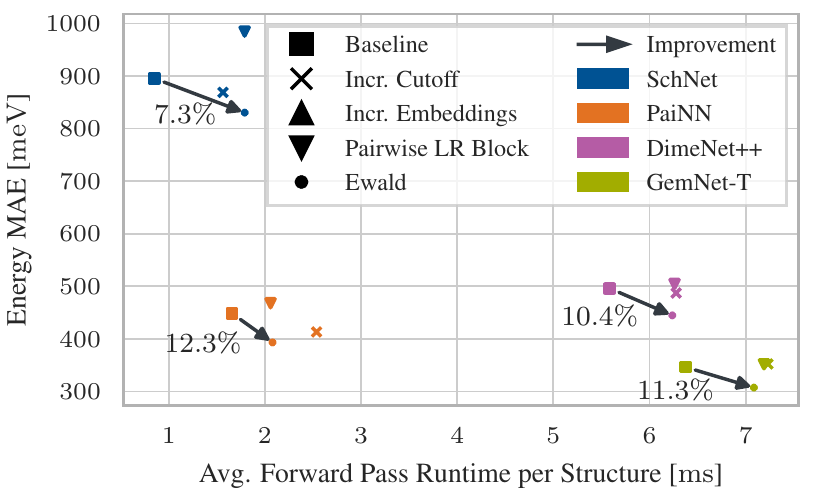}
\vskip -0.05in
\caption{Cost vs.\ energy MAE on OC20, averaged across test splits. Ewald MP consistently improves all models and significantly outperforms the two baseline settings featuring an increased global cutoff as well as a standard pairwise (here: SchNet) block with increased distance cutoff in place of the Ewald LR block.}
\label{fig:pareto_oc20}
\end{minipage}%
\end{figure*}

\paragraph{Ewald modifications.}
We aim to upgrade each baseline architecture in a minimal way such that the Ewald additions can be studied in isolation. All models except DimeNet$^{++}$ update atom embeddings in each interaction block using a skip connection. We simply add long-range message passing using $M_i^{\text{lr}}$ from \cref{eq:lrmessage,eq:s} as another contribution,
\begin{align}
\begin{split}
&h_i^{(l+1)} = \frac{1}{\sqrt{2}} \left[h_i^{(l)} + f^{\text{sr}}_{\text{upd}}\left(M_i^{\text{sr}}\right) \right] \\
&\downmapsto\\
&h_i^{(l+1)} = \frac{1}{\sqrt{3}} \left[h_i^{(l)} + f^{\text{sr}}_{\text{upd}}\left(M_i^{\text{sr}}\right) + f^{\text{lr}}_{\text{upd}}\left(M_i^{\text{lr}}\right)\right],
\end{split}
\label{eq:ewaldmod}
\end{align}
where the short-range message sum $M_i^{\text{sr}}$ and update function $f^{\text{sr}}_{\text{upd}}$ is specific to the short-range model, while $f^{\text{lr}}_{\text{upd}}$ is of the same kind across models (architecture details in \cref{app:architecture}). For PaiNN, which also incorporates equivariant vector embeddings, we only upgrade messages between its scalar embeddings. As DimeNet$^{++}$ does not pass messages between atom embeddings by default, we add them to the model with a default embedding block taken from GemNet-T and update them purely by long-range message passing with a skip connection, $h_i^{(l+1)} = \frac{1}{\sqrt{2}} \left[h_i^{(l)} + f_{\text{upd}}\left(M_i^{\text{lr}}\right)\right]$. We concatenate these atom embeddings with the atom-aggregated DimeNet$^{++}$ edge embeddings along the feature dimension, prior to the atom-wise dense layer of each output block. Any remaining differences among our Ewald modifications are on the level of hyperparameters. Aiming for Pareto-efficient baseline improvements, we choose numbers of included $k$-vectors adequate to the cost of each baseline (see \cref{app:ewaldhparams}).

\paragraph{Comparison studies.}
We test whether using Ewald message passing alongside our baselines leads to significant and robust improvements of the energy mean absolute errors (EMAEs) on OC20 and OE62. We also measure the average runtime per structure (details in \cref{app:runtime}) for every model configuration and dataset to study how the benefits of Ewald message passing compare against its runtime overhead. Moreover, to establish a baseline for the long-range improvements achievable by a mere change in hyperparameters, we repeat the EMAE and runtime measurements in a configuration without Ewald MP, but with increased settings for the distance cutoff $c_x$ and maximum number of message-passing neighbors $N_{\text{max}}$ (details in \ref{app:comp_hyperparams}). An adequate comparison ideally means increasing these settings to a point that roughly matches the Ewald modification in cost. This is always possible for the periodic structures of OC20. In the case of OE62, a setting of $c_x = \SI{12}{\angstrom}$ and $N_{\text{max}}=100$ usually covers the entire molecule, meaning that no significant cost increase can be achieved beyond this point. We furthermore compare the Ewald improvements against a test setting where the Ewald block, $f^{\text{lr}}_{\text{upd}}\left(M_i^{\text{lr}}\right)$ in \cref{eq:ewaldmod}, is instead replaced by regular pairwise message passing with increased distance cutoff (similarly to related work discussed in \cref{sec:rw}). Concretely, we substitute it by a SchNet interaction block. In contrast to the increased-cutoff variant from before, we now extend the cutoff for this block \emph{only}. As a final check, we study whether the benefits of Ewald MP are significant beyond what follows just from an increase in free parameters. On OE62, we measure the EMAE and runtime of baseline variants with more atom embedding channels to recover the parameter count of Ewald MP. For OC20, we point to the extensive hyperparameter studies by \citet{gasteiger_how_2022}, which suggest that changing embedding sizes by the small amounts relevant to this work has negligible impact on OC20 EMAEs.\par

First off, our comparison studies on OE62 (\cref{fig:pareto_oe62}) and OC20 (\cref{fig:pareto_oc20}) underscore the primary role of short-range model choice in the GNN design landscape (consider, e.g., how DimeNet$^{++}$ is pareto-dominated on OC20). At the same time, the consistency of accuracy gains through Ewald MP strongly encourages its use alongside any particular short-range model. The associated runtime overheads are at least comparable to, if not smaller than the runtime differences between baselines. Meanwhile, the average Ewald improvements across all models (and all test splits in the case of OC20) are \SI{16.1}{\percent} for OE62, and \SI{10.3}{\percent} for OC20. Ewald MP achieves a better cost-performance tradeoff than the large cutoff setting for all tested instances on OC20. Interestingly, the large cutoff setting on OE62 has a very comparable tradeoff to Ewald MP for the two baselines which use edge embeddings, but has a detrimental effect for the two models that only use atom embeddings. Quite the opposite happens on OE62 if the Ewald block is replaced by a SchNet block with increased distance cutoff: it is effective for PaiNN and SchNet, but achieves no improvement at all on the better DimeNet++ and GemNet-T baseline models. Its cost cannot be further increased as the added SchNet block already approaches full graph connectivity. On OC20, it always has a neutral or detrimental effect. In contrast to these two previous strategies, improvements through Ewald MP appear \emph{robust} -- no detrimental effect is observed in \emph{any} tested model configuration. Finally, matching the free parameter count of Ewald MP with larger atom embeddings shows a negligible or even detrimental effect. Beyond these empirical comparisons, we note in general that pairwise message passing on a fully-connected graph (like our SchNet LR block setting on OE62) inevitably scales quadratically in structural size. Ewald MP, on the other hand, can propagate long-range information subquadratically while also providing a physically principled inductive bias (cf. \cref{sec:EwaldMP}).\par
Our results inform various design recommendations for future use. Ewald MP appears to be more efficient on OC20: similar relative improvements are consistently achieved at about half the relative cost. This hints at periodic structures as the more attractive domain of application.
Given how we integrate Ewald MP with a short-range GNN, it is relatively cheaper alongside models featuring few but compute-heavy message passing blocks (in our case, DimeNet$^{++}$ and GemNet-T). In such cases, adding a long-range message sum to each block results in less relative overhead. Besides energies, we also test for improvements of force MAEs on OC20, which are consistent but small. This is to be expected, since the frequency truncation scheme of long-range MP removes the fast-changing high-frequency contributions to the potential energy surface, limiting its impact on the energy gradients (forces). Further comments and numerical results are found in \cref{app:forces,app:results}.

\begin{figure}[t!]
\includegraphics[width=\linewidth]{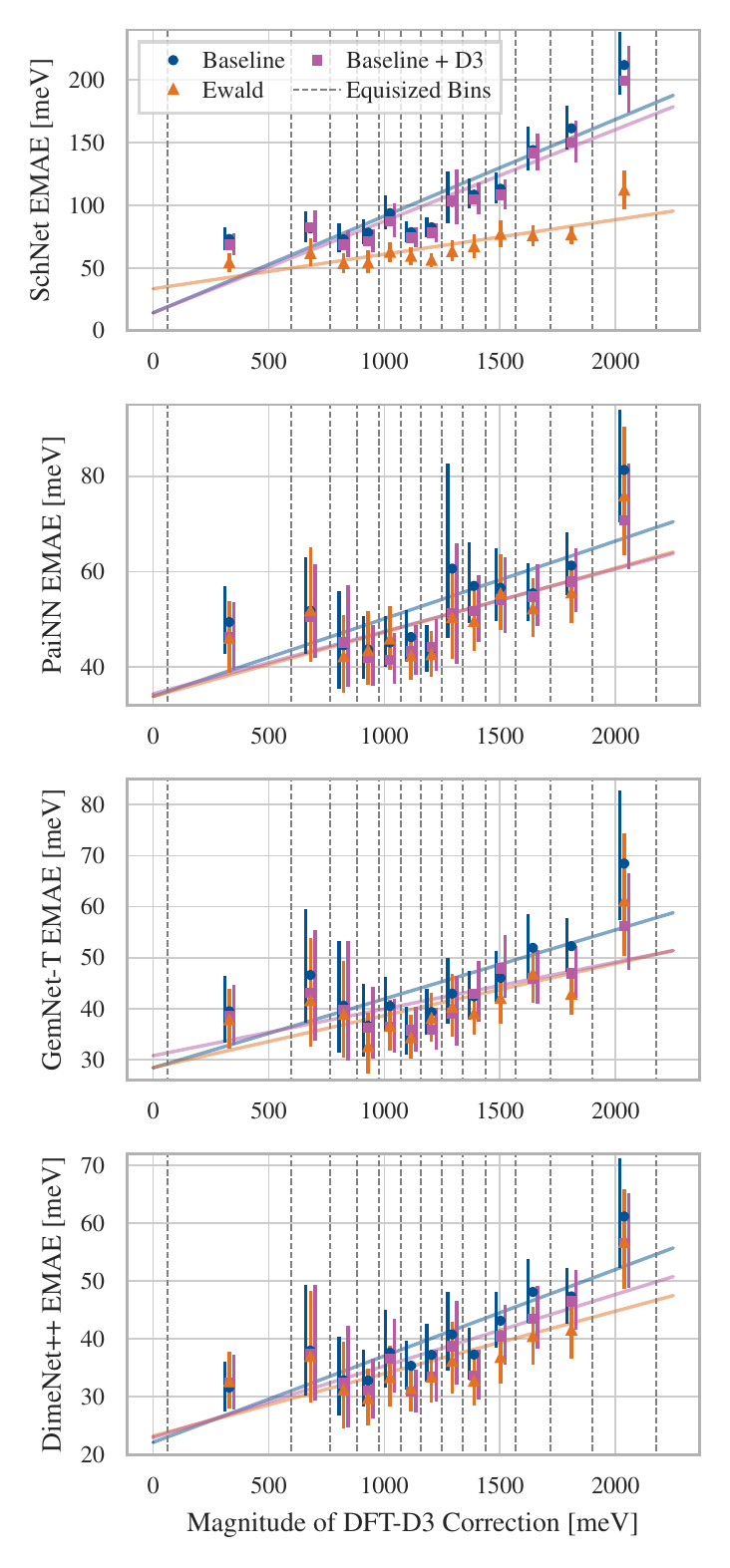}
\vskip -0.05in
\caption{Energy MAE on \texttt{OE62-val} structures binned according to their DFT-D3 long-range term. Ewald MP has an outsize impact for molecules with larger long-range components, and recovers or surpasses a ``cheating'' baseline that just adds the D3 ground truth to the prediction (``Baseline + D3'').}
\label{fig:lranalysis_short}
\end{figure}

\paragraph{Impact on long-range interactions.} To test our hypothesis that Ewald MP learns long-range effects in particular, we leverage the fact that the energy targets on OE62 contain an additive term, DFT-D3 \cite{dft-d3}, to correct for the long-ranged London-dispersion interaction. This allows us to study the long-range component in isolation.
To evaluate the DFT-D3 term, we use the coefficient set recommended for Becke-Johnson damping and the PBE0 DFT functional, as obtained from \citet{bjdampingparams}. We proceed by assigning all 6000 molecules in the \texttt{OE62-val} dataset to one of 15 equally-sized bins according to the value of its DFT-D3 correction. We disregard the two outermost bins as they contain strong outliers. In each bin, we record the statistics (bootstrapped means and \SI{68}{\percent} confidence intervals with 10000 resamples) of the energy MAE for the baseline and Ewald model variants. 
\cref{fig:lranalysis_short} shows that Ewald message passing gives an outsize relative improvement for structures with high long-range energy contributions. As an additional check, we collect the same statistics as previously for a model without Ewald modifications that is trained while the DFT-D3 correction is added explicitly to the model output. This provides a ``cheating'' model that only needs to learn the short-range component, while the long-range component is perfectly reconstructed. Ewald message passing keeps up with this perfect baseline across all models, and even outperforms it for SchNet and DimeNet$^{++}$. This analysis shows that Ewald MP is able to recover the effects of a dispersion correction on OE62 without any need for hand-crafting.

\section{Conclusion}

In this work, we show how the analogy between sums of GNN messages and physical interactions leads to Ewald message passing, a general message-passing framework based on Ewald summation. Being computationally efficient and model-agnostic, it brings value as a plug-and-play addition to many existing GNN models. Our results demonstrate robust improvements in energy targets across models and datasets. They furthermore indicate an outsize impact on structures with high long-range energy contributions. We encourage using Ewald MP especially for large or periodic structures containing a diverse set of atoms.


\section*{Acknowledgements}
We greatly thank Abhishek Das for his swift help in evaluating our models on the \texttt{OC20-test} split, as well as the anonymous reviewers for their instructive suggestions.\par
This work is funded by the Munich Quantum Valley by the Federal Ministry of Education and Research (BMBF), from funds of the Hightech Agenda Bayern Plus, and the Federal Ministry of Education and Research and the Free State of Bavaria under the Excellence Strategy of the Federal Government and the Länder.


%

\bibliography{bibliography}
\bibliographystyle{icml2023}
\clearpage


\appendix
\section{Appendix}
\begin{figure*}[t!]
\includegraphics[width=\linewidth]{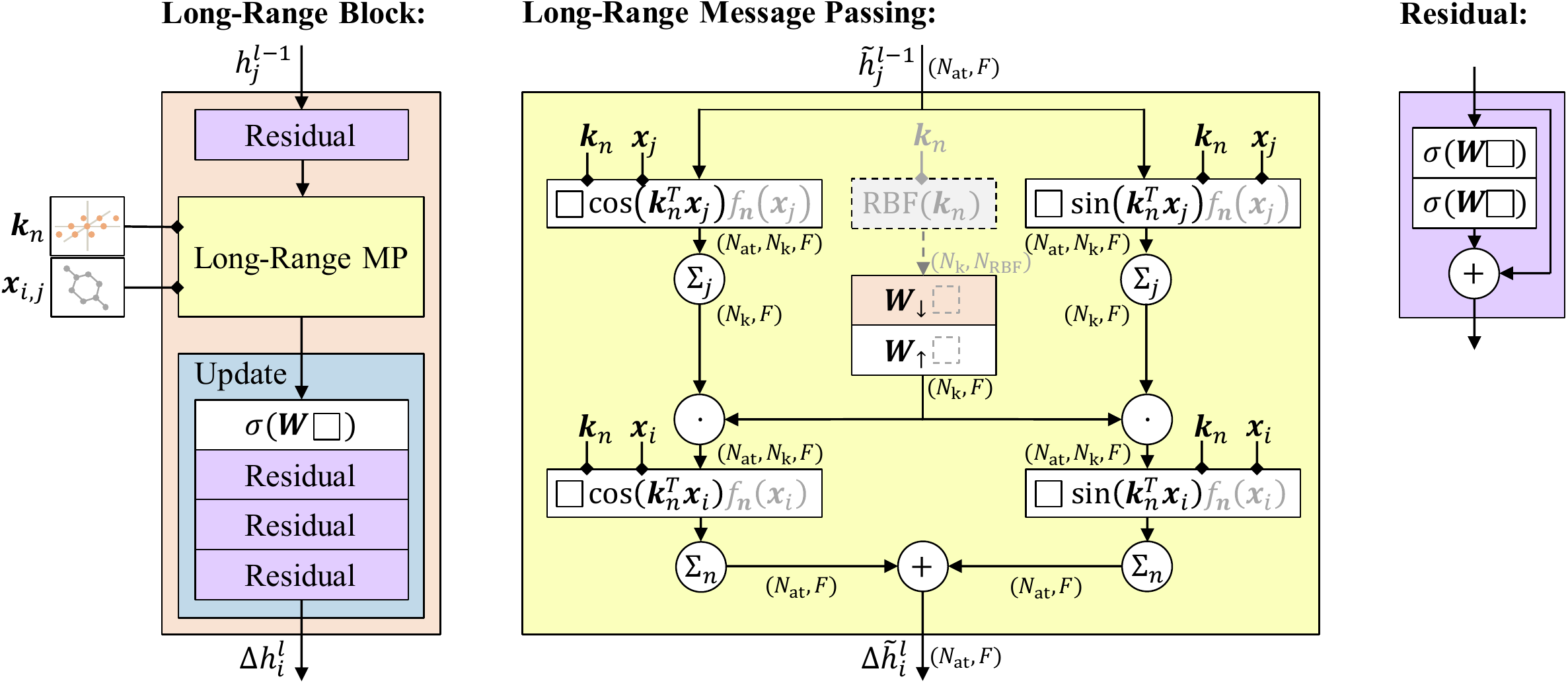}
\caption{The architecture of the long-range part of Ewald message passing. The long-range block on the left corresponds to the expression $f^{\text{lr}}_{\text{upd}}\left(M_i^{\text{lr}}\right)$. $\Box$ denotes the input of a layer, encircled operations act component-wise. Triangular arrowheads define information flow through layers, diamond arrowheads denote parametric dependence. Annotated in brackets are the shapes of tensors handled throughout various parts of the computation. The steps marked by gray and dashed lines apply only to the aperiodic case.}
\label{fig:ewald_architecture}
\end{figure*}

\subsection{Architecture and Implementation Details}
\label{app:architecture}
\cref{fig:ewald_architecture} shows detailed computing steps used to obtain the expression $f^{\text{lr}}_{\text{upd}}\left(M_i^{\text{lr}}\right)$ (as given in \cref{eq:ewaldmod}), which we add to the GNN-specific short-range embedding update as part of a skip connection during every interaction step (\cref{eq:ewaldmod}). The part highlighted in yellow computes the long-range sum $M_i^{\text{lr}}$ that we discuss in \cref{sec:EwaldMP}. The index $i$ denotes the target atom, while the source index $j$ runs over all atoms in the structure (or the atoms of a single supercell in the case of periodic boundary conditions).

\textbf{Real and imaginary parts.} While we use complex exponential notation to simplify the conceptual part of this work, our implementation handles the real and imaginary parts of the structure factor individually. \cref{eq:s} gives $s_\vk = \sum_{j \in \mathcal{S}} h_j \cos(\vk^T\vx_j), \, \text{Im}(s_\vk) = - \sum_{j \in \mathcal{S}} h_j \sin(\vk^T\vx_j)$. Plugging this into \cref{eq:lrmessage} and considering the real and imaginary parts of $M_i^{\text{lr}}$, we obtain

\begin{equation}
\begin{split}
\text{Re}(M_i^{\text{lr}}) &= \sum_{n} \cos(\vk_n^T\vx_i) \sum_{j \in \mathcal{S}} h_j \cos(\vk_n^T\vx_j) \hat \Phi^{\text{lr}}(\vk_n)\\
&+ \sum_{n} \sin(\vk_n^T\vx_i) \sum_{j \in \mathcal{S}} h_j \sin(\vk_n^T\vx_j) \hat \Phi^{\text{lr}}(\vk_n),\\
\text{Im}(M_i^{\text{lr}}) &= 0.
\label{eq:realimag}
\end{split}
\end{equation}

This prescription leads to the steps outlined in \cref{fig:ewald_architecture}. In the aperiodic case in which we apply radial filtering with a direct cutoff, the index $n$ enumerates all frequencies within the cutoff, $\{\vk_n \in \Lambda', \, \lVert \vk_n \rVert \leq c_k\}$, and we have $\hat \Phi(\vk_n) \equiv \hat \Phi(\lVert \vk_n \rVert)$. In the periodic case, we drop the radial symmetry of $\hat \Phi$ as well as the notion of a direct cutoff, and instead identify $n \equiv \lambda' \, \in I$, with the index set $I$ enumerating reciprocal lattice sites close to the origin as defined in \cref{sec:EwaldMP}. To obtain the above result (particularly $\text{Im}(M_i^{\text{lr}}) = 0$), we implicitly use that the Fourier filter coefficients $\hat \Phi^{\text{lr}}(\vk_n)$ are real-valued, yet the real-valuedness of the \emph{position-space} potential $\Phi^{\text{lr}}$ alone would \emph{only} guarantee $\hat \Phi^{\text{lr}}(-\vk_n) = \hat \Phi^{\text{lr}}(\vk_n)^*$, with $*$ denoting complex conjugation. The real-valuedness of all coefficients $\hat \Phi^{\text{lr}}(\vk_n)$ follows from the additional point symmetry constraint that we impose on the Fourier space filters. In the aperiodic filtering case, the point symmetry follows directly from radial symmetry, $\hat \Phi^{\text{lr}}(\lVert -\vk_n \rVert) = \hat \Phi^{\text{lr}}(\lVert \vk_n \rVert)$. In the periodic case, we enforce the symmetry via $\hat \Phi_{\bm{\lambda'}} \equiv \hat \Phi_{\bm{-\lambda'}}$ for all $\bm \lambda' \in I$. The reason for this constraint is that an $\sR^3$ lattice spanned by an arbitrary basis always has point symmetry around the origin – the model should therefore give rise to the same output on two structures that differ only by a point reflection of their respective PBC frames while their atom positions are equal.

\textbf{Damping values.} In the aperiodic case, all trigonometric functions in \cref{eq:realimag} are multiplied by a damping value $f_n(\vx)$, which depends on the position $\vx$ as well as the frequency $\vk_n$, and which can differ depending on the chosen discretization scheme for the structure factor (\cref{app:structurefactor}). In our tested case of discretization by voxel averaging with voxel resolution $\Delta$, we have
\begin{equation}
f_n(\vx) = \prod_{c \, \in \, \{x,y,z\}}{\text{sinc}\left(\frac{k_n^c x^c \Delta}{2} \right)},
\end{equation}
where the superscript $c$ enumerates vector components.

\textbf{Implementation of filtering schemes.}
Apart from damping values, the second, implementation-related difference between the periodic and aperiodic case comes from our choice of Fourier-space filtering scheme (\cref{sec:EwaldMP}). In the aperiodic case, the matrix product $\mW_\uparrow \mW_\downarrow$ gives a weight matrix $\mW$, which we apply to the set of radial basis function values to get the filter values $\{\hat \Phi^{\text{lr}}(\lVert \vk_n \rVert), \, \vk_n \, \in \, \Lambda', \, \vk_n \leq c_k\}$ (as discussed in the aperiodic filtering paragraph of \cref{sec:EwaldMP}). In the periodic case, we do not apply the product $\mW_\uparrow \mW_\downarrow$ directly to a vector, but rather evaluate and store the resulting matrix given our current pair of weight matrices $(\mW_\uparrow, \mW_\downarrow)$. We identify the transpose $(\mW_\uparrow \mW_\downarrow)^T$ of this matrix with the filter coefficients $\{\hat \Phi_{\bm{\lambda'}} = \hat \Phi^{\text{lr}}(\vk_{\bm \lambda'}) = \hat \Phi^{\text{lr}}(\vk_n), \, \bm \lambda' \in I\}$ (recall our identification of $n$ with the multi-index $\bm \lambda'$). In both the periodic and aperiodic cases, we end up with a matrix of shape $(N_{\text{k}}, F)$ containing $F$-dimensional stacks of filter values $\hat \Phi^{\text{lr}}(\vk_n)$, that we multiply component-wise with the structure factor according to the prescription in \cref{eq:realimag}. We use the downprojection $\mW_\downarrow$ onto a bottleneck layer of dimension $N_{\downarrow}$ (with shared weight values among all interaction blocks $l$) to encourage generalization, as inspired by \citet{gasteiger_gemnet_2021}. Hence, $\mW_\uparrow$ has shape $(F, N_{\downarrow})$, while $\mW_\downarrow$ has shape $(N_{\downarrow}, N_{\text{k}})$ (periodic case) or $(N_{\downarrow}, N_{\text{RBF}})$ (aperiodic case). We set $N_{\downarrow}=16$ for GemNet-T and $N_{\downarrow}=8$ otherwise.

\textbf{Update function.}
As the update function $f^{\text{lr}}_{\text{upd}}$, we use among all models a dense layer followed by $N_{\text{hidden}}$ residual layers, as specified by \citet{gasteiger_gemnet_2021} in the atomic update block of the GemNet architecture. While the authors report using $N_{\text{hidden}} = 2$ residual blocks, we set $N_{\text{hidden}} = 3$ for all models except PaiNN, where we set $N_{\text{hidden}} = 0$, opting for less relative overhead compared the inexpensive short-range blocks of the PaiNN architecture.

\subsection{Training and Hyperparameters \label{app:training}}
\textbf{Baseline hyperparameters,} unless explicitly listed here, are the same as the values used for the baseline models in \citep{chanussot_open_2021}. All differences from these original OCP hyperparameter settings either serve a fairer assessment of Ewald modifications by improving baseline performance (especially on the energy MAE metric primarily addressed by Ewald message passing), or were introduced to achieve more practical training times for our experimental setup. Except for the new hyperparameters entering with Ewald message passing, we use exactly the same model settings for the baseline and Ewald versions. On OE62, we use embedding sizes of 512 for the PaiNN and SchNet models and 256 for the GemNet-T and DimeNet$^{++}$ models. On OC20, we had to reduce the DimeNet$^{++}$ filter size to 192 due to memory constraints. All baselines use distance cutoff values of \SI{6}{\angstrom} and neighbor thresholds $N_{\text{max}}$ of 50. Our SchNet and PaiNN models use 4 interaction blocks, DimeNet$^{++}$ and GemNet-T use 3 blocks.\par

\textbf{Training hyperparameters on OE62.} We find on OE62 that smaller batch sizes improve performance for all baselines except SchNet. Hence, we use OE62 batch sizes of 64 (SchNet) and 8 (else). All models have an initial learning rate of $1\cdot10^{-4}$ except for GemNet-T ($5\cdot10^{-4}$). As in the PaiNN reference \citep{painn_schutt}, we use the Adam optimizer with weight decay $\lambda = 0.01$, along with a plateau scheduler (patience 10 and decay factor 0.5). For DimeNet$^{++}$ and GemNet-T, we use a milestone scheduler with a warmup period of 250000 steps (warmup factor 0.2) and decay factor $0.1$ at 750000, 1125000 and 1500000 steps. For SchNet, we also use a milestone scheduler, with 30000 warmup steps (warmup factor 0.2) and decay factor $0.1$ at 60000, 120000 and 180000 steps. Training on OE62 converges after 200 maximum epochs for PaiNN, 250 epochs for SchNet and DimeNet$^{++}$, and 300 epochs for GemNet-T.\par

\textbf{Training hyperparameters on OC20.} To reduce training times, we increase batch sizes on OC20. GemNet-T and DimeNet$^{++}$ use batch size 16, PaiNN batch size 48 and SchNet batch size 64. We use a plateau scheduler with patience 3 and factor 0.8 for PaiNN, training for 25 maximum epochs. GemNet-T uses the same equal plateau scheduler. We train SchNet with a milestone scheduler (90000 warmup steps with warmup factor 0.2, milestones with factor 0.1 at 150000, 250000, 310000 steps) for 25 epochs, and DimeNet$^{++}$ with 250000 warmup steps (warmup factor 0.2) and factor 0.1 milestones at 750000, 1125000 and 1500000 steps for 15 epochs. In the loss function, we use force multipliers of $33.3$ for SchNet, $100$ for PaiNN, $100$ for GemNet-T, and $16.7$ for DimeNet$^{++}$.

\textbf{Loss function.} Like \citet{chanussot_open_2021}, we use a linear combination of energy and force mean absolute errors. Using the index $\mathcal S$ to run over the $|\mathcal D|$ atom structures in the set $\mathcal D$ (which becomes a mini-batch in practice), as well as $i$ to run over the  $|\mathcal S|$ individual atoms in a structure $\mathcal S$, our loss function has the form
\begin{equation}
\begin{gathered}
\frac{1}{|\mathcal D|}\sum_{\mathcal S \in \mathcal D}\left|E^{(\mathcal S)}-E^{(\mathcal S)}_{\text{t}}\right| \\
+\frac{\lambda}{|\mathcal D|} \sum_{\mathcal S \in \mathcal D} \frac{1}{|\mathcal S|}\left \lVert\bm F_i^{(\mathcal S)}- \bm F^{(\mathcal S)}_{i; \text{t}}\right \rVert_p,
\end{gathered}
\end{equation}
with the predicted/target truth energies $E^{(\mathcal S)}$/$E^{(\mathcal S)}_{\text{t}}$, the predicted/target forces $\bm F_i^{(\mathcal S)}$/$\bm F^{(\mathcal S)}_{i; \text{DFT}}$, the force multiplier $\lambda \in \sR_+$ defining the desired tradeoff between energy and force accuracy, and $\lVert \cdot \rVert_p$ denoting the Euclidean $p$-norm. According with model literature, we use $p=1$ for SchNet and DimeNet$^{++}$, and $p=2$ for PaiNN and GemNet-T. On OE62, where there are only energy targets, we set $\lambda = 0$.

\subsection{Ewald Hyperparameters and Ablation Studies} \label{app:ewaldhparams}

\textbf{Aperiodic case.}
Ewald message passing introduces two main new hyperparameters in the aperiodic case: the frequency cutoff $c_k$ and the voxel resolution $\Delta$. An additional source of hyperparameters is the parametrization of the frequency filters. In the case of our implementation, the number of Gaussian radial basis functions $N_{\text{RBF}}$ needs to be fixed. In order to achieve pareto-efficient improvements through Ewald message passing, we adapt these hyperparameters to the comparative cost of each short-range baseline GNN. Our main experiment settings are listed in \cref{tab:ewald_hyperparams}.

\begin{table}[h!]
\begin{center}
\begin{tabular}{rlllll}
          & $c_k [\si{\angstrom^{-1}}]$ & $\Delta [\si{\angstrom^{-1}}]$ & $N_{\text{RBF}}$ &  &  \\ \cline{1-4}
SchNet    & 0.4                         & 0.2                            & 48    &  &  \\
PaiNN     & 0.6                         & 0.2                            & 128   &  &  \\
DimeNet$^{++}$ & 0.8                         & 0.2                            & 128   &  &  \\
GemNet-T  & 1.0                         & 0.2                            & 128   &  & 
\end{tabular}
\caption{Aperiodic Ewald hyperparameter settings.}
\label{tab:ewald_hyperparams}
\end{center}
\end{table}

\textbf{Periodic case.}
In the periodic case, the main hyperparameters of Ewald message passing are the numbers $N_x, N_y, N_z$ which define the frequencies included in the long-range sum. In principle, they can all be varied independently, but in our work we tie them into one independent effective hyperparameter by the prodedure explained below.\par
Recall that we keep the amount of included frequencies fixed across structures in our periodic filtering setting, which leads to better performance empirically. This rules out simply including all reciprocal lattice frequencies within a fixed cutoff volume, as their total number would vary across input structures with differing reciprocal lattice supercells. Still, we impose that the range of included frequencies does not extend out to significantly further distance in any of the three reciprocal lattice directions, just as it would be the case with a spherical cutoff volume. In our setting, this is not possible for an arbitrary input structure, as the set of included reciprocal lattice indices $I$ (defined in \cref{sec:EwaldMP}, and fixed via our hyperparameters) is mapped onto different points of Fourier space for any structure via $\vk_{\bm \lambda'} = \lambda'_1 \vw_1 + \lambda'_2 \vw_2 + \lambda'_3 \vw_3$. However, we can make sure that an even Fourier space distance is covered by included frequencies along all three lattice directions for the \emph{average} structure. For instance, if the reciprocal lattice points are twice as dense on average in one direction as compared to another, we double $N_i$ for this direction to have the same average Fourier space distance covered by frequencies along both axes. This single distance effectively acts like an ``average cutoff'', while the number of included frequencies stays fixed for all structures. On OC20, we observe that the average reciprocal space unit cell is roughly 3 times narrower along the z direction (the surface normal) than it is along the $x$ and $y$ directions. Applying this hyperparameter tying constraint and accounting for different baseline model runtimes leads to our main settings listed in \cref{tab:ewald_hyperparams_periodic}.

\begin{table}[h!]
\begin{center}
\begin{tabular}{rlllll}
          & $N_x$ & $N_y$ & $N_z$ &  &  \\ \cline{1-4}
SchNet    & 1                         & 1                            & 3    &  &  \\
PaiNN     & 1                         & 1                            & 3   &  &  \\
DimeNet$^{++}$ & 2                         & 2                           & 5   &  &  \\
GemNet-T  & 2                         & 2                            & 5   &  & 
\end{tabular}
\caption{Periodic Ewald hyperparameter settings.}
\label{tab:ewald_hyperparams_periodic}
\end{center}
\end{table}

\textbf{Additional hyperparameters.}
Additional hyperparameters introduced by Ewald message passing in both the periodic and aperiodic case are the bottleneck dimension $N_\downarrow$ and the amount of residual layers $N_{\text{hidden}}$ used as part of the update function (cf. \cref{app:architecture}).

\textbf{Ablation studies: aperiodic case.}
We test OE62 EMAEs of the DimeNet$^{++}$-Ewald model with alternative settings for $c_k$ and $\Delta$ to probe the robustness of our approach in hyperparameter space. We obtain EMAEs of \SI{46.9}{\milli\electronvolt} for ($c_k =\SI{0.6}{\angstrom^{-1}}$, $\Delta = \SI{0.2}{\angstrom^{-1}}$), \SI{47.2}{\milli\electronvolt} for ($c_k =\SI{0.4}{\angstrom^{-1}}$, $\Delta = \SI{0.1}{\angstrom^{-1}}$), and \SI{47.3}{\milli\electronvolt} for ($c_k =\SI{1.6}{\angstrom^{-1}}$, $\Delta = \SI{0.4}{\angstrom^{-1}}$), all of which is a consistent improvement over the \SI{51.2}{\milli\electronvolt} baseline MAE.\par
We also test similar variations of the SchNet model, obtaining EMAEs of
\SI{83.0}{\milli\electronvolt} for ($c_k =\SI{0.4}{\angstrom^{-1}}$, $\Delta = \SI{0.4}{\angstrom^{-1}}$), \SI{85.9}{\milli\electronvolt} for ($c_k =\SI{0.2}{\angstrom^{-1}}$, $\Delta = \SI{0.2}{\angstrom^{-1}}$), and \SI{82.8}{\milli\electronvolt} for ($c_k =\SI{0.8}{\angstrom^{-1}}$, $\Delta = \SI{0.4}{\angstrom^{-1}}$) which are again very close to our main setting results and strongly improve the \SI{133.5}{\milli\electronvolt} baseline MAE.

\textbf{Ablation studies: periodic case.}
We test OC20 EMAEs (averaged across all four test splits) of the SchNet-Ewald model with different settings for $(N_x, \, N_y, \, N_z)$, probing whether our hyperparameter tying constraint (see above) is a sensitive choice. We obtain EMAEs of \SI{836}{\milli\electronvolt} $(1, 1, 1)$ and \SI{828}{\milli\electronvolt} $(1, 1, 2)$. Again, we find that our improvements over the \SI{895}{\milli\electronvolt} baseline MAE are robust to different settings.\par

We do not use radial filters in the PBC setting as we find them to perform worse empirically. We train our GemNet-T-Ewald in a time-reduced setting of 30 maximum message-passing neighbors and for 12 epochs on OC20-2M. We record EMAEs on the four \texttt{OC20-val} splits. Our standard filtering approach with $(N_x, \, N_y, \, N_z) = (2,2,5)$ has 137 filter weights (due to mirror symmetrization as mentioned in \cref{sec:EwaldMP}). We compare this to an Ewald variant using radial filtering with 128 Gaussian radial basis functions. Furthermore, we test another filtering approach which uses 3D Gaussian basis functions $\mathcal G(\Sigma, \mu_i)$ with covariance matrices $\Sigma = \text{diag}(\Delta_x^2, \Delta_y^2, \Delta_z^2) \in \sR^{3x3}$ centered on the points $\mu_i \in \Lambda'_c$ of a cuboid lattice $\Lambda'_c = \{\lambda'_x \Delta_x (1,0,0)^T + \lambda'_y \Delta_y (0,1,0)^T +\lambda'_z \Delta_z (0,0,1)^T \mid \bm \lambda' \in I\}$, with $I = \bigtimes_{j=x,y,z} \{-N_j, \dots, N_j\} \subset \sZ^3$. We test $(N_x, N_y, N_z) = (3,3,6)$ and $(\Delta_x, \Delta_y, \Delta_z) = (0.333, 0.333, 0.167)\si{\angstrom}^{-1}$. As seen above, all variants improve the baseline, but the radial and cuboid filtering variants generalize worse on the OOD-both split, in which both the catalyst and adsorbate have out-of-distribution composition (\cref{tab:filterabl}).

\begin{table}[]
\begin{centering}
\begin{tabular}{rllll}
Setting    & Baseline & Standard & Radial & Cuboid \\ \hline
ID  & 305      & 263      & 264 & 278   \\
OOD & 455      & 416      & 447 & 436  
\end{tabular}
\end{centering}
\caption{EMAEs [\si{\milli\electronvolt}] of filtering variants on OC20.}
\label{tab:filterabl}
\end{table}

\subsection{Hyperparameters for Comparison Studies}
\label{app:comp_hyperparams}
\textbf{Increased cutoff setting.}
For the spatially infinite structures of OC20, we assume the amount of message passing neighbors for any given cutoff roughly satisfies $N \sim c_x^3$ and scale it up accordingly. Given starting values of $c_x = \SI{6}{\angstrom}$ and $N_{\text{max}}=50$ for all baselines, we find the modified OC20 settings $(\SI{6.25}{\angstrom}, 55)$ for GemNet-T and DimeNet$^{++}$ $(\SI{7}{\angstrom}, 80)$ for PaiNN and $(\SI{8.5}{\angstrom}, 140)$ for SchNet which roughly match the respective Ewald model variants in cost. On OE62, we double the neighbor threshold from 50 to 100 (exceeding the total size of most structures), use an increased $\SI{9}{\angstrom}$ cutoff for GemNet-T and a $\SI{12}{\angstrom}$ cutoff for all other baselines. On OE62, we double the neighbor threshold from 50 to 100 (exceeding the total size of most structures), use an increased $\SI{8}{\angstrom}$ cutoff for GemNet-T and $\SI{12}{\angstrom}$ for all other baselines.

\textbf{Increased embedding size setting.}
In our OE62 increased embedding size study, we increase sizes to 1536 for SchNet, 576 for PaiNN, 384 for DimeNet$^{++}$ and 320 for GemNet-T, which approximately matches the free parameter counts of the Ewald variants (\cref{tab:freeparams}).
\begin{table}[]
\setlength{\tabcolsep}{4pt}
\begin{centering}
\begin{tabular}{rllll}
Setting    & SchNet & PaiNN & DimeNet$^{++}$ & GemNet-T \\ \hline
Ewald  & 12.2M      & 15.7M      & 4.8M & 16.1M   \\
Emb. & 14.4M      & 15.7M      & 5.4M & 15.8M 
\end{tabular}
\end{centering}
\caption{Parameter counts (in million parameters) for the Ewald and increased-embedding variants on OE62.}
\label{tab:freeparams}
\end{table}

\textbf{Pairwise long-range block setting.}
We replace the Ewald block ($f^{\text{lr}}_{\text{upd}}\left(M_i^{\text{lr}}\right)$ in \cref{eq:ewaldmod}) by a standard SchNet block with 200 Gaussians and 256 filters. On OE62, cost saturates at 12 Å and 100 maximum neighbors, where the molecular graph approaches full connectivity. On OC20, long-ranged SchNet blocks cannot compete with the Ewald cost: the Ewald overhead is already matched at 6.25 Å for PaiNN, 6.75 Å for SchNet and DimeNet++, and 7.25 Å for GemNet-T, where we scale up the neighbor threshold as in the increased cutoff setting. Increasing distance cutoffs on OC20 is more costly due to the higher atomic density.

\subsection{Relationship between $\hat M_\vk^{\text{lr}}$ and $\hat \Phi^{\text{lr}}(\lVert \cdot \rVert)$}
\label{app:relationship}
To derive the desired identity as done in standard Ewald summation, we start with the Fourier coefficients $\hat M_\vk^{\text{lr}}$,
\begin{equation}
\hat M_\vk^{\text{lr}} = \frac{1}{\Omega} \int_{\Omega(\mathcal S)} M^{\text{lr}}(\vx) \exp(-i\vk^T\vx) d^3\vx,
\end{equation}
where $\Omega(\mathcal{S}) \subset \sR^3$ denotes a single supercell parallelepiped. We now plug in \cref{eq:periodic} for $M^{\text{lr}}(\vx)$, resulting in
\begin{equation}
\begin{split}
\hat M_\vk^{\text{lr}} = \frac{1}{\Omega} \int_{\Omega(\mathcal S)} \sum_{\vt \in \Lambda} \sum_{j \in \mathcal{S}} \, &h_j \Phi^{\text{lr}}\left(\lVert \vx - \vx_j - \vt \rVert\right)\\
\cdot &\exp(-i\vk^T \vx) d^3\vx.
\end{split}
\end{equation}

\emph{Assuming} the lattice sum over $\Lambda$ is absolutely convergent, we can exchange it with the supercell integral such that 
\begin{equation}
\begin{split}
   &\int_{\Omega(\mathcal S)} \sum_{\vt \in \Lambda} \Phi^{\text{lr}}\left(\lVert \vy_j - \vt \rVert\right) \exp(-i\vk^T \vy_j) d^3\vy_j\\
    = &\sum_{\vt \in \Lambda} \int_{\Omega(\mathcal S)}\Phi^{\text{lr}}\left(\lVert \vy_j - \vt \rVert\right) \exp(-i\vk^T \vy_j) d^3\vy_j\\
    = &\int_{\sR^3} \Phi^{\text{lr}}\left(\lVert \vy_j \rVert\right) \exp(-i\vk^T \vy_j) d^3\vy_j
\end{split}
\end{equation}
holds (since the supercells form a tessellation of $\sR^3$), where we have set $\vy_j \vcentcolon = \vx - \vx_j$. We now identify the obtained integral over $\sR^3$ as the Fourier transform $\hat \Phi^{\text{lr}}(\lVert \vk \rVert)$:
\begin{equation}
\begin{split}
&\hat M_\vk^{\text{lr}} = \frac{1}{\Omega} \sum_{j \in \mathcal S}  h_j \left[\int_{\sR^3} \Phi^{\text{lr}}(\lVert \vy_j \rVert) \exp(-i \vk^T \vy_j) d^3\vy_j \right]\\
\cdot &\exp(-i\vk^T \vx_j)
= \frac{1}{\Omega} \sum_{j \in \mathcal S} h_j \exp(-i\vk^T \vx_j) \hat \Phi^{\text{lr}}(\lVert \vk \rVert).
\end{split}
\end{equation}

\subsection{Discretization of the Structure Factor \label{app:structurefactor}}
There is no unique approach to discretization and we believe our application leaves future room for more sophisticated schemes, e.g., approaches based on wavelet theory \citep{mallat_wavelets}. To aid such efforts, we first present a quite general framework for discretizing the structure factor. Afterwards, we outline our concrete implementation and show how it arises as a special case. Assume a finite set of $N_{\text{k}}$ square-integrable, complex-valued basis functions $\{f_i \in L^2(D) \mid 1 \leq i \leq N_{\text{k}}\}$ on a bounded domain $D$ containing the cutoff domain $\mathcal{B}_c = \{\vk \in \sR^3 \mid k \leq c_k\}$, i.e. $\mathcal{B}_c \subset D \subset \sR^3$. Further assume orthonormality $\langle \, f_j, f_k \rangle_{L^2} = \int_{D}f_j(\vk)^*f_k(\vk) d^3\vk = \delta_{jk}$, where $^*$ denotes complex conjugation. The structure factor $s(\vk)$ can be projected onto this basis set, resulting in a discrete approximate representation (\cref{eq:discretization}) in terms of basis coefficients $\{s_i \mid  1 \leq i \leq N_{\text{k}}\}$:
\begin{equation}
    s(\vk) \approx \sum_{i=1}^{N_{\text{k}}} \langle \, f_i, s \rangle f_i(\vk) =\vcentcolon \sum_{i=1}^{N_{\text{k}}} s_i f_i(\vk)
    \label{eq:discretization}
\end{equation}
Integrating these scalar products numerically would require knowing $s(\vk)$ at high resolution. Instead, \cref{eq:sfcoefficients} takes advantage of the appearing Fourier transform:
\begin{align}
\begin{split}
    s_i &= \langle \, f_i, s \rangle =  \sum_{b \in \mathcal{S}} h_b \int_{D} f_i(\vk)^* \exp(-i\vk\vx_b) d^3\vk \\
    &= \sum_{b \in \mathcal{S}} h_b \widehat{[f_i^*]}(\vx_b) = \sum_{b \in \mathcal{S}} h_b \hat{f}_i^*(-\vx_b).
    \label{eq:sfcoefficients}
\end{split}
\end{align}
Note the analogy to $s_{\vk}$, with $\hat{f}_i^*(-\vx_b)$ replacing $\exp(-i\vk\vx_b)$. We illustrate this with two examples.\par

\textit{(a) Auxiliary supercell.} A standard literature approach \citep{supercell_finite_size_extrapolation} is to approximate an aperiodic structure as part of a supercell that is assumed finite, but sufficiently large that spurious interactions can be ignored. This reintroduces a reciprocal lattice $\Lambda'$ on which discrete $s_{\vk_i} \vcentcolon= s(\vk_i)$ can be evaluated. Due to the defining property of the delta distribution, $\langle \, \delta(\cdot - \vk_i), s \rangle = s(\vk_i)$, this is a special (i.e., limiting) case of our formalism featuring a distributional 'basis' and Fourier transform,
\begin{align}
\begin{split}
\{&f^{\text{SC}}_i(\vk) = \delta(\vk-\vk_i) \mid \vk_i \in \Lambda' \cap \mathcal{B}_c\},\\
\label{eq:deltafourier}
&\hat{f}_i^{\text{SC}*}(-\vx_b) = \exp(-i\vk_i\vx_b),
\end{split}
\end{align}

As seen above from \cref{eq:deltafourier}, the definition of coefficient $s_i$ in \cref{eq:sfcoefficients} indeed reduces to the PBC definition of $s_{\vk}$ in this special case.\par

\textit{(b) Voxel basis.} In our implementation of aperiodic Ewald message passing, we discretize $s(\vk)$ by taking voxel averages $s_i$ on the cutoff-covering portion $\Lambda'_\Delta \cap \mathcal{B}_c$ of a 3D voxel grid $\Lambda'_\Delta = \{\lambda'_1 \Delta (1,0,0)^T + \lambda'_2 \Delta (0,1,0)^T +\lambda'_3 \Delta (0,0,1)^T \mid \lambda' \in \sZ^3\}$ with voxel sidelength $\Delta$. The elements $\vk_i \in \Lambda'_\Delta \cap \mathcal{B}_c$ are the voxel center locations. In our general framework, this corresponds to a basis set
\begin{align}
\begin{split}
\{&f^{\text{V}}_i = \Delta^{-3} \mathcal{X}_{\vk_i} \mid \vk_i \in \Lambda'_\Delta \cap \mathcal{B}_c\},\\
\label{eq:voxelfourier}
&\hat{f}_i^{\text{V}*}(-\vx_b) = \exp(-i\vk_i\vx_b) \prod_{j\in\{xyz\}}{\text{sinc}\left(\frac{k_i^j x^j_b \Delta}{2} \right)},
\end{split}
\end{align}
where $\mathcal{X}_{\vk_i}$ takes on a value of 1 the $\vk_i$-centered cube $\prod_{j\in \{x,y,z\}}{\left[k_i^j - \frac{\Delta}{2}, k_i^j + \frac{\Delta}{2}\right]}$ and 0 otherwise. The hyperparameter $\Delta$ controls the discretization resolution. Taking the scalar product,
\begin{equation}
    \langle \, f^{\text{V}}_i, s \rangle = \Delta^{-3} \int_{k_i^x-\frac{\Delta}{2}}^{k_i^x+\frac{\Delta}{2}} \int_{k_i^y-\frac{\Delta}{2}}^{k_i^y+\frac{\Delta}{2}} \int_{k_i^z-\frac{\Delta}{2}}^{k_i^z+\frac{\Delta}{2}}s(\vk)d^3\vk,
\end{equation}
computes the average value of $s(\vk)$ inside the voxel cube around $\vk_i$. Using the Fourier transform result from \cref{eq:voxelfourier}, we see that $s_i$ only differs from the PBC case $s_{\vk}$ by an additional sinc damping function. Taking $\Delta \rightarrow 0$ recovers the continuum limit: consistently, $\text{sinc}\left(\frac{x_b^j\Delta}{2}\right)$ converges pointwise to $1$ in this limit.\par
So far, we have outlined how to represent the structure factor by discrete basis coefficients $s_i$. To compute long-range messages in our voxel implementation, we proceed as follows. The continuous filter $\hat \Phi(k)$ is assigned the same type of voxel representation $\sum_{i=1}^{N_{\text{k}}} \hat\Phi_i f_i^{\text{V}}$ in terms of coefficients $\hat\Phi_i = \langle \, f_i^{\text{V}}, \hat \Phi \rangle$. These coefficients are obtained by reading out the learned radial basis combinations $\hat \Phi(k) = \mW \hat \Psi(k)$ on the discrete grid of voxel centerpoints $\Lambda'_\Delta \cap \mathcal{B}_c$. Now, the complete expression for the message sum assumes the approximate form
\begin{align}
\begin{split}
    M_a^{\text{lr}} &= \int_D \exp(i\vk\vx_a) \left[\sum_{i=1}^{N_{\text{k}}} s_i \hat \Phi_i f_i^{\text{V}}(\vk)\right] d^3\vk \\
    = &\sum_{i=1}^{N_{\text{k}}} \hat f_i^{\text{V}}(-\vx_a) \left[\sum_{b} h_b \hat f_i^{\text{V}*}(-\vx_b)\right] \hat \Phi_i,
\end{split}
\end{align}
where in the last step we substituted in the expression for $s_i$ from \cref{eq:sfcoefficients}. Note the analogy to the PBC message passing formula \cref{eq:lrmessage} with $\exp(i\vk\vx_a)$ and $\exp(-i\vk\vx_b)$ replaced by $f_i^{\text{V}}(-\vx_a)$ and $f_i^{\text{V}*}(-\vx_b)$, respectively. As these differ only by an additional sinc damping factor for the voxel case, our periodic and aperiodic implementations of Ewald message passing use just marginally different code.


\subsection{Impact on Force Predictions}
\label{app:forces}
Ewald message passing is primarily a correction scheme for energy rather than force targets. The force on atom $i$ is the negative energy gradient $\bm{F}(\vx_i) = -\nabla_i E(\vx_1, \dots \vx_N)$. Due to the wavelength truncation used in \cref{eq:lrmessage}, Ewald message passing can only contribute terms to the energy prediction which vary more slowly in space than the scale of its inverse frequency cutoff. To see this quantitatively, consider the Fourier series ansatz for the function $V_i(\vx) \vcentcolon= E(\vx_1, \dots \vx_{i-1}, \vx, \vx_{i+1}, \vx_N)|_{\vx_1, \vx_{i-1}, \dots, \vx_{i+1}, \vx_N}$, reading $V_i(\vx) = \sum_{\vk \in \Lambda'} \hat V_\vk \exp{(i \vk^T \vx)}$. In these terms,
\begin{equation}
    \bm{F}(\vx_i) = -\nabla V(\vx_i) = -\sum_{\vk \in \Lambda'} i \vk \hat V_\vk \exp{(i \vk^T \vx_i)}.
    \label{eq:force_energy_gradient}
\end{equation}
Contributions at low frequencies $\lVert \vk \rVert \leq c_k$ are therefore being suppressed compared to the gradient contributions of higher frequencies. Hence, we hypothesize that the \emph{short}-range message sums of the model take over a major part of the force prediction.

\subsection{Runtime Measurements}
\label{app:runtime}
We run all models on \texttt{Nvidia A100} GPUs and evaluate the runtimes of all model configurations in one session and on the same machine. After 50 warmup batches, we average the runtime per structure over 500 batches and repeat this measurement three times. Afterwards, we take the minimum of these three measurements.

\subsection{Further Long-Range Binning Results}
As a supplement to our analysis in the main body of this work, we provide additional binning results similar to \cref{fig:lranalysis_short}. They give additional insights about the correlation of various structural features with Ewald improvements.

\begin{figure}[h!]
\includegraphics[width=\linewidth]{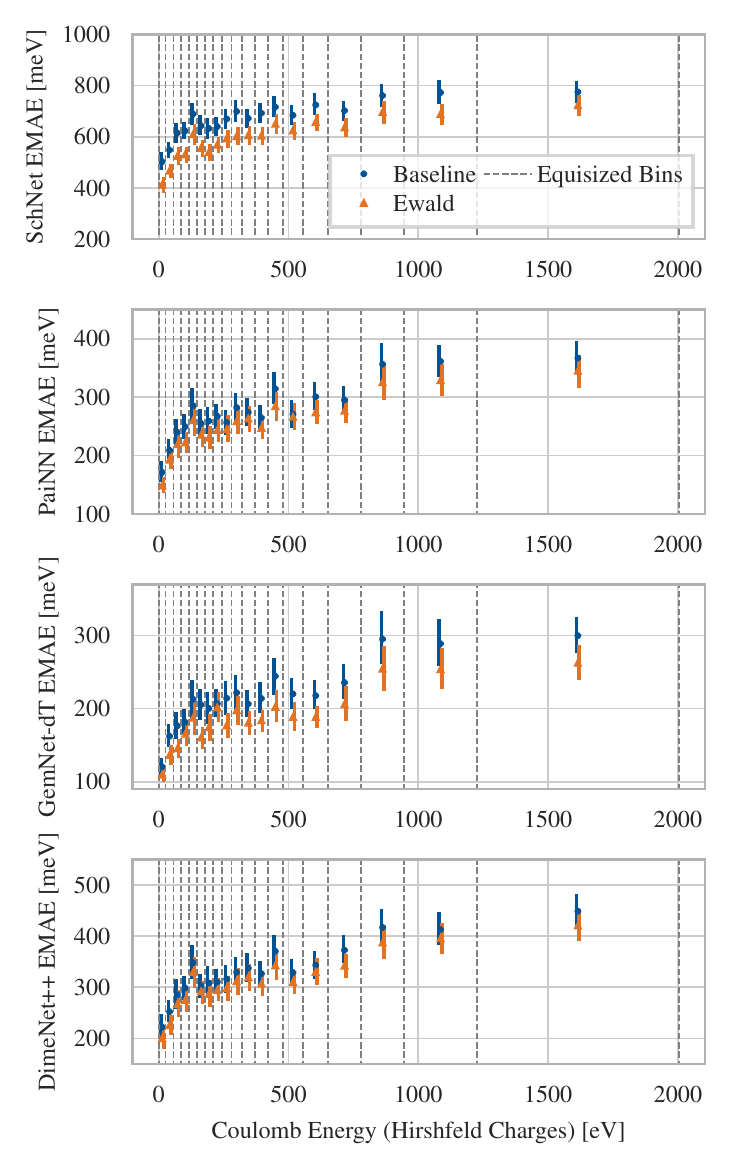}
\vskip -0.08in
\caption{Energy MAE on \texttt{OC20-IS2RE-val-id} structures 
binned according to electrostatic energy. Ewald MP has \emph{no} outsize impact on structures with a high electrostatic energy contribution.}
\vskip -0.1in
\label{fig:lr_plots_oc20}
\end{figure}

\textbf{Further OE62 analyses.}
We group structures by two structural size metrics: the maximum distance between any pair of atoms, and the standard deviation $\sigma$ along the principal axis for the atomic point cloud. Furthermore, we bin according to electrostatic energy \cref{eq:esenergy}, obtained using DFT-Hirshfeld atomic partial charges from the OE62 data.

\textbf{Further OC20 analyses.}
On OC20, no obvious size metric exists due to the periodic boundary conditions. However, we can still bin according to electrostatic energy as for OE62 on the \texttt{IS2RE-val-id} split of OC20, for which Bader partial charges are provided on all structures.

\textbf{Discussion.}
Our additional analyses inform the following conclusions and usage recommendations:
\begin{itemize}
    \item{Ewald MP appears to address dispersion effects. We hope our method might also learn such effects for reference targets going beyond DFT-D3 in the cost hierarchy, such as energies obtained from a nonlocal DFT functional.}
    \item{We observe no increased relative improvement for higher electrostatic energy, only a constant correction across all bins. Apparently, standard GNNs can already learn electrostatics fairly well on our data. It could be valuable to repeat this study on upcoming benchmarks in fields like interface or colloid chemistry.}
    \item{We expect the usefulness of Ewald MP to grow with structural size due to its favorable scaling. However, our binning plots suggest that structural size is no direct proxy for expected model error or Ewald improvement.}
\end{itemize}

\subsection{OE62 Dataset Preprocessing}
\label{app:preprocessing}
As the raw OE62 DFT energies have a large average offset from zero, we compute least-squares-fitted targets for all structures $\mathcal{S}$ of the form $E^{(\mathcal{S})}_{\text{t}} = E^{(\mathcal{S})}_{\text{raw}} - \sum_{Z} C_Z N^{(\mathcal{S})}_Z - C_0$, where $E^{(\mathcal{S})}_{\text{t}}$/$E^{(\mathcal{S})}_{\text{raw}}$ are the preprocessed target / raw DFT energies, $Z \in \sN$ enumerates atom types with type-specific regression coefficients $C_Z \in \sR$ and bias coefficient $C_0$, and $N^{(\mathcal{S})}_Z \in \sN_0$ is the number of atoms belonging to type $Z$ in the structure $\mathcal{S}$. The model only learns the residuals of this fit, which have close-to-zero mean by construction. We partition the data in ca. 50000 structures for \texttt{OE62-train}, and ca. 6000 structures for each of \texttt{OE62-val} and \texttt{OE62-test}.

\subsection{Numerical Energy, Force and Runtime Results}
\label{app:results}
\cref{table:results_oc20,table:results_oe62} provide the numerical OC20 and OE62 results on which we base our visualizations in \cref{sec:exp}.
\newpage
\begin{sidewaysfigure*}[h!]
\includegraphics[width=\linewidth]{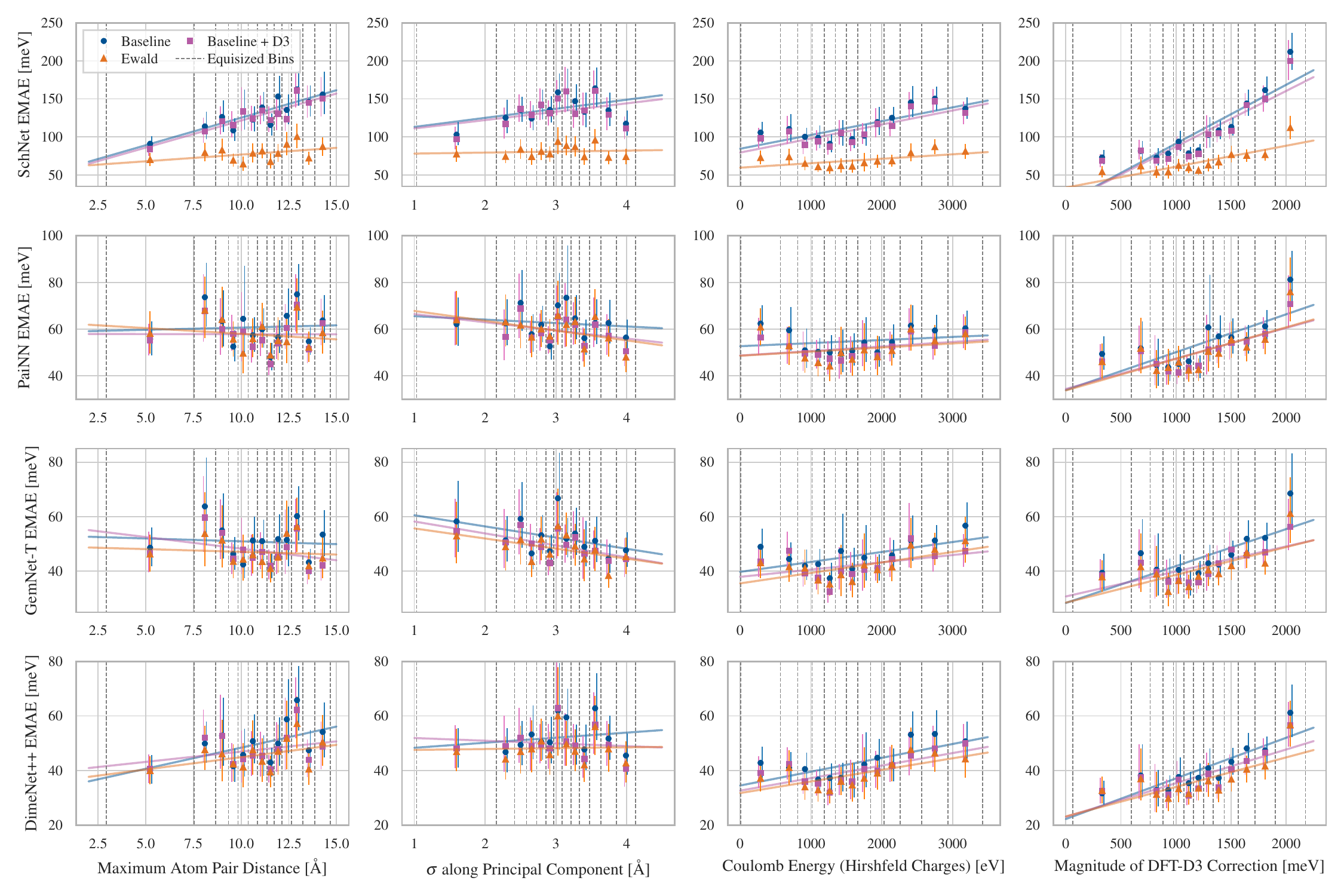}
\caption{Binning of the energy EMAE on \texttt{OE62-val} by two structural size metrics, electrostatic and dispersion energy. Only dispersion is a good predictor for Ewald improvement.}
\label{fig:lr_plots_full}
\end{sidewaysfigure*}

\begin{table*}[t]
\begin{center}
\caption{Energy MAEs and runtimes per input structure on OE62.}
\vspace{10pt}
\begin{tabular}{cccccccccc}
          &            & \multicolumn{2}{c}{\texttt{OE62-val}}                                         & \multicolumn{2}{c}{\texttt{OE62-test}}                                        & \multicolumn{2}{c}{Forward Pass}                                                & \multicolumn{2}{c}{Forward \& Backward Pass}                                               \\ \cline{3-10} 
          &            & MAE                                     & Rel.                       & MAE                                     & Rel.                       & Runtime                                                & Rel.                         & Runtime                                                & Rel.                         \\
Model     & Variant  & $\SI{}{\milli\electronvolt} \downarrow$ & $\SI{}{\percent} \uparrow$ & $\SI{}{\milli\electronvolt} \downarrow$ & $\SI{}{\percent} \uparrow$ & $\SI{}{\milli\second\per\text{struct.}} \downarrow$ & $\SI{}{\percent} \downarrow$ & $\SI{}{\milli\second\per\text{struct.}} \downarrow$ & $\SI{}{\percent} \downarrow$ \\ \hline
SchNet    & Baseline   & 133.5                                   & -                          & 131.3                                   & -                          & 0.13                                                & -                            & 0.28                                                & -                            \\
          &  Embeddings & 144.7                                   & -8.4                       & 136.7                                   & -4.1                       & 0.14                                                & 15.2                         & 0.33                                                & 17.8                         \\
          &  Cutoff     & 257.4                                   & -92.8                      & 254.8                                   & -94.1                      & 0.14                                                & 13.6                         & 0.31                                                & 11.6                         \\
          &  SchNet-LR     & 86.6                                   & 35.1                      & 89.2                                   & 32.1                      & 0.32                                                & 156.0                         & 0.75                                                & 171.7                         \\
          & Ewald      & \textbf{79.2}                                    & \textbf{40.7}                       & \textbf{81.1}                                    & \textbf{38.2}                       & 0.70                                               & 461.6                        & 1.03                                                & 271.4                        \\
          \hline
PaiNN     & Baseline   & 61.4                                    & -                          & 63                                      & -                          & 1.52                                                & -                            & 3.16                                                & -                            \\
          &  Embeddings & 63.5                                    & -3.4                       & 63.1                                    & -0.2                       & 1.54	&1.4	&3.28	&3.8                         \\
          &  Cutoff     &     65.1                     &       -6.0                     &      64.4                          &   -2.2                     &          1.84	& 20.9	& 3.91	& 23.6                             \\
          &  SchNet-LR     &     58.3                     &       5.1                     &      58.2                          &   7.7                     &          1.84	& 20.7	& 4.21	& 33.1                             \\
          & Ewald      & \textbf{57.9}                                    & \textbf{5.7}                        & \textbf{59.7}                                    & \textbf{5.2}                        & 2.29	& 50.5	& 4.57	& 44.4                       \\
          \hline
DimeNet$^{++}$ & Baseline   & 51.2                                    & -                          & 53.8                                    & -                          & 1.99                                               & -                            & 4.26                                               & -                            \\
          &  Embeddings & 50.4                                    & 1.6                        & 53.4                                    & 0.7                        & 2.25	&12.9	&4.93	&15.8                         \\
          &  Cutoff     & 48.3                                    & 5.7                        & \textbf{48.1}                                    & \textbf{10.6}                       & 2.68	&34.7	&6.10	&43.4                         \\ 
          &  SchNet-LR     &     51.4                     &       -0.5                     &      54.4                          &   -1.1                     &          2.37	& 19.0	& 4.73	& 11.2                             \\
          & Ewald      & \textbf{46.5}                                    & \textbf{9.2}                        & \textbf{48.1}                                    & \textbf{10.6}                       & 2.70	&35.5	&5.93	&39.5                         \\
          \hline
GemNet-T  & Baseline   & 51.5                                    & -                          & 53.1                                    & -                          & 3.07                                                & -                            & 6.96                                                & -                            \\
          &  Embeddings & 52.7                                    & -2.3                       & 53.9                                    & -1.5                       & 3.11	&1.5	&6.98	&0.4                          \\
          &  Cutoff     & 47.8                                    &       7.2                 &        47.7                       &          10.2              &  4.02	& 31.2	&8.88	&27.7   \\
          &  SchNet-LR     &     51.2                     &       0.6                     &      52.8                          &   0.5                     &          3.32	& 8.3	& 7.73	& 11.1  \\
          & Ewald      & \textbf{47.4}                                    & \textbf{8.0}                        & \textbf{47.5}                                    & \textbf{10.5}                       & 4.05	& 32.0	& 8.86	& 27.4                         
\end{tabular}
\label{table:results_oe62}

\end{center}
\end{table*}

\begin{table*}
\begin{center}
\setlength{\tabcolsep}{3pt}
\caption{Energy and force mean absolute errors, as well as runtimes per input structure, averaged across all four splits of \texttt{OC20-test}.}
\vspace{10pt}
\begin{tabular}{cccccccccc}
          &           & \multicolumn{2}{c}{Energy}                                           & \multicolumn{2}{c}{Forces}                                                        & \multicolumn{2}{c}{Forward Pass}                                                & \multicolumn{2}{c}{Forward \& Backward Pass}                                               \\ \cline{3-10} 
          &           & MAE                                     & Rel.                       & MAE                                                  & Rel.                       & Runtime                                                & Rel.                         & Runtime                                                & Rel.                         \\
Model     & Variant & $\SI{}{\milli\electronvolt} \downarrow$ & $\SI{}{\percent} \uparrow$ & $\SI{}{\milli\electronvolt\per\angstrom} \downarrow$ & $\SI{}{\percent} \uparrow$ & $\SI{}{\milli\second\per\text{struct.}} \downarrow$ & $\SI{}{\percent} \downarrow$ & $\SI{}{\milli\second\per\text{struct.}} \downarrow$ & $\SI{}{\percent} \downarrow$ \\ \hline
SchNet    & Baseline  & 895                                     & -                          & 61.1                                                 & -                          & 0.85                                                & -                            & 1.83                                                & -                            \\
          & Cutoff    & 869                                     & 3.0                        & 60.3                                                 & 1.3                        & 1.57                                                & 84.2                         & 3.92	& 114.2                        \\
          & SchNet-LR    & 984                                     & -9.9                        & 65.3                                                 & -6.8                        & 1.79                                                & 110.6                         & 5.24	& 186.0                        \\
          & Ewald     & \textbf{830}                                     & \textbf{7.3}                        & \textbf{56.7}                                                 & \textbf{7.2}                        & 1.79                                                & 110.7                        & 4.68	& 155.3                        \\
          \hline
PaiNN     & Baseline  & 448                                     & -                          & 42.0                                                 & -                          & 1.66                                                & -                            & 3.44                                               & -                            \\
          & Cutoff    & 413                                     & 7.9                        & 40.9                                                 & 2.6                        & 2.54                                                & 53.1                         & 5.06	& 46.9                         \\
          & SchNet-LR    & 468                                     & -4.3                        & 43.3                                                 & -3.2                        & 2.06                                                & 24.2                         & 5.24	& 44.3                         \\
          & Ewald     & \textbf{393}                                     & \textbf{12.3}                       & \textbf{40.5}                                                 & \textbf{3.5}                        & 2.08                                                & 25.4                         & 4.24 &	23.1                         \\
          \hline
DimeNet$^{++}$ & Baseline  & 496                                     & -                          & 47.1                                                 & -                          & 5.58                                                & -                            & 15.68                                               & -                            \\
          &  Cutoff    & 487                                     & 1.8                        & 46.4                                                 & 1.4                        & 6.28                                                & 12.4                         & 17.23	& 9.9                          \\ 
          &  SchNet-LR    & 504                                     & -1.6                        & 48.3                                                 & -2.6                        & 6.26                                                & 12.1                         & 16.14	& 2.9                          \\ 
          & Ewald     & \textbf{445}                                     & \textbf{10.4}                       & \textbf{45.5}                                                 & \textbf{3.3}                        & 6.24                                                & 11.7                         & 17.00	& 8.4                          \\
          \hline
GemNet-dT  & Baseline  & 346                                     & -                          & 29.9                                                 & -                          & 6.37                                                & -                            & 14.77                                                & -                            \\
          &  Cutoff    & 352                                     & -1.7                        & 30.2                                                 & -1.0                       & 7.23                                                & 13.5                         & 16.40	& 11.0                     \\
          &  SchNet-LR    & 351                                     & -1.5                        & 29.8                                                 & 0.3                       & 7.19                                                & 12.1                         & 16.35	& 10.7                     \\
          & Ewald     & \textbf{307}                                     & \textbf{11.3}                       & \textbf{29.3}                                                 & \textbf{1.8}                        & 7.09                                                & 11.2                         & 16.23	& 9.9                          
\end{tabular}
\label{table:results_oc20}
\end{center}
\end{table*}

\end{document}